\documentclass[conference]{IEEEtran}
\IEEEoverridecommandlockouts
\usepackage{cite}
\usepackage{amsmath,amssymb,amsfonts}
\usepackage{algorithmic}
\usepackage{graphicx}
\usepackage{textcomp}
\usepackage{xcolor}
\usepackage{hyperref}
\usepackage{xcolor}
\usepackage{braket}
\usepackage{comment}
\usepackage[normalem]{ulem}
\usepackage[caption=false]{subfig}

\usepackage{booktabs} 
\usepackage{tabularx} 

\def\BibTeX{{\rm B\kern-.05em{\sc i\kern-.025em b}\kern-.08em
    T\kern-.1667em\lower.7ex\hbox{E}\kern-.125emX}}

\makeatletter
\newcommand{\linebreakand}{%
  \end{@IEEEauthorhalign}
  \hfill\mbox{}\par
  \mbox{}\hfill\begin{@IEEEauthorhalign}
}
\makeatother

\author{
\IEEEauthorblockN{Abdallah Aaraba\textsuperscript{\textdagger*}, Soumaya Cherkaoui\textsuperscript{\textsection}, Ola Ahmad\textsuperscript{\textparagraph}, Jean-Frédéric Laprade\textsuperscript{\textdaggerdbl}, \\ Olivier Nahman-Lévesque\textsuperscript{\textdaggerdbl}, Alexis Vieloszynski\textsuperscript{\textdagger}, and Shengrui Wang\textsuperscript{\textdagger}}\\

\IEEEauthorblockA{\textsuperscript{\textdagger}Departement of Computer Science, Université de Sherbrooke, Sherbrooke, Canada}
\IEEEauthorblockA{\textsuperscript{\textsection}Department of Computer and Software Engineering, Polytechnique Montréal, Montréal, Canada}
\IEEEauthorblockA{\textsuperscript{\textparagraph}Thales Digital Solutions, Montréal, Canada}
\IEEEauthorblockA{\textsuperscript{\textdaggerdbl}Institut Quantique, Université de Sherbrooke, Sherbrooke, Canada}\\
\IEEEauthorblockA{\textsuperscript{*}Email: abdallah.aaraba@usherbrooke.ca}
}

\title{QuaCK-TSF: Quantum-Classical Kernelized Time Series Forecasting \thanks{© 2024 IEEE. Personal use of this material is permitted. Permission from IEEE must be obtained for all other uses, in any current or future media, including reprinting/republishing this material for advertising or promotional purposes, creating new collective works, for resale or redistribution to servers or lists, or reuse of any copyrighted component of this work in other works.}}

\begin{document}

\maketitle

\begin{abstract}
Forecasting in probabilistic time series is a complex endeavor that extends beyond predicting future values to also quantifying the uncertainty inherent in these predictions. Gaussian process regression stands out as a Bayesian machine learning technique adept at addressing this multifaceted challenge. 
This paper introduces a novel approach that blends the robustness of this Bayesian technique with the nuanced insights provided by the kernel perspective on quantum models, aimed at advancing quantum kernelized probabilistic forecasting. We incorporate a quantum feature map inspired by Ising interactions and demonstrate its effectiveness in capturing the temporal dependencies critical for precise forecasting. The optimization of our model's hyperparameters circumvents the need for computationally intensive gradient descent by employing gradient-free Bayesian optimization. Comparative benchmarks against established classical kernel models are provided, affirming that our quantum-enhanced approach achieves competitive performance.

\end{abstract}

\begin{IEEEkeywords}
Quantum machine learning, Quantum kernels, Probabilistic time series forecasting, Gaussian processes regression.
\end{IEEEkeywords}

\section{Introduction}
Quantum computing is poised to revolutionize both scientific research and industrial applications \cite{arute2019quantum, bravyi2022future, 
bayerstadler2021industry, bova2021commercial},
driven by rapid hardware advancements \cite{bornens2023variational, madsen2022quantum, blais2021circuit} and significant algorithmic developments aimed at harnessing quantum advantage beyond basic demonstrations \cite{kim2023evidence, bravyi2020quantum, rainjonneau2023quantum, piatkowski2022towards}. In particular, quantum machine learning (QML), which integrates quantum computing with machine learning, is anticipated to be among the early beneficiaries of these advancements \cite{liu2021rigorous}. Nevertheless, the size and coherence of the quantum circuits needed to attain a regime where quantum computers can tackle real world problems 
exceeds the capabilities of the present noisy intermediate-scale quantum (NISQ) technology\cite{preskill2018quantum}.

Recently, quantum kernel methods have emerged as a promising approach to leveraging the capabilities of current NISQ devices when viewing QML models \cite{havlivcek2019supervised}. The value of this perspective is rooted in the compatibility of quantum models with the extensive range of classical kernel theory tools, which can significantly streamline the optimization process for quantum models \cite{schuld2021supervised, scholkopf2001generalized}.
These methods involve mapping
data into the Hilbert space associated to a system of qubits 
using 
a quantum feature map that facilitates the calculation of a kernel matrix through the pairwise inner product of data points. This matrix can subsequently be utilized in conventional methods such as support vector machines or kernel ridge regression \cite{burges1998tutorial,vovk2013kernel}.
The underlying premise is that quantum Hilbert space data encoding can enhance the feature map with quantum-exclusive resources, offering benefits over classical maps. Such advantages have been demonstrated for specific datasets \cite{huang2022quantum, liu2021rigorous}.

Time series forecasting (TSF) represents a domain ripe for the transformative impacts of quantum computing. This field encompasses critical activities such as financial forecasting \cite{gui2014financial}, where the potential speed-up offered by quantum computing could lead to markedly faster predictions, conferring a significant competitive edge \cite{thakkar2024improved}. The implications of such advancements are profound, particularly in the high-stakes realm of finance where rapid and accurate predictions are paramount. Additionally, the role of forecasting is becoming increasingly pivotal in security contexts. This importance is underscored by the recent development of methods that connect QML advancements with time series anomaly detection challenges \cite{kalfon2023successive, liu2018quantum}, showcasing the potential of QML in addressing complex forecasting tasks.

Quantifying uncertainty in time series forecasts is a critical component of the prediction process \cite{aaraba2024fr}.
Consequently, probabilistic TSF techniques have garnered considerable attention, especially in areas where managing uncertainty is vital, such as anomaly detection. Gaussian process regression (GPR) stands out as a simple yet effective probabilistic TSF  method \cite{rasmussen2006gaussian}. Based on Bayesian inference, this approach adeptly captures 
uncertainty by presenting the forecast of the next value in a time series as a posterior 
probability distribution. This distribution not only provides an expected value for the next point in the series (i.e., the mean) but
also quantifies 
the confidence in this prediction through the variance, offering a comprehensive view of future expectations and their reliability.

In recent research, the focus has been predominantly on exploring quantum versions of traditional kernel machines, notably the support vector machine \cite{rebentrost2014quantum, havlivcek2019supervised}, showcasing their potential in leveraging quantum computing capabilities. Despite this interest, the development and exploration of quantum probabilistic kernel methods within the context of noisy intermediate-scale quantum (NISQ) technologies remain relatively underexplored \cite{rapp2024quantum}. This gap is particularly evident in the field of probabilistic TSF, where the application of quantum-enhanced probabilistic kernel methods has barely been addressed in scholarly works.

In pursuit of advancing the field of quantum kernelized probabilistic TSF, we introduce QuaCK-TSF: Quantum-Classical Kernelized Time Series Forecasting. Our model endeavors to predict the subsequent value of a time series $\mathbf{s} = (x_{t_1}, x_{t_2}, ..., x_{t_T})$ at time step $t_{T+1}$, with each $x_{t_l} \in \mathbb{R}$ and time $t_l \in \mathbb{R}^+$. Leveraging the GPR framework, we assert that for a given kernel function $\kappa$ and a training dataset $\mathcal{D}$ derived from the series $\mathbf{s}$, the posterior distribution $p(x_{t_{T+1}} | \mathcal{D})$ can be modeled as a Gaussian distribution $\mathcal{N}\left(\bar{x}_{t_{T+1}}, \sigma_{t_{T+1}}^2\right)$, where $\bar{x}_{t_{T+1}}$ represents the mean prediction, and $\sigma_{t_{T+1}}^2$ signifies the prediction's variance.

We construct a portal to quantum-enhanced forecasting by employing a quantum kernel that is conceptualized as the state fidelity overlap between quantum states \cite{schuld2021supervised}. Our model employs a variant of the instantaneous quantum polynomial-time (IQP) feature map \cite{havlivcek2019supervised}, similar to the approach in \cite{dai2022quantum}, as detailed in subsection \ref{subsubsec:IQP-feature-map}.
The choice of the IQP feature map is motivated by its conjectured resistance to classical simulation \cite{havlivcek2019supervised} and its ability to capture temporal relationships through qubit interactions, akin to Ising models. By encoding temporal data points as qubit states, the feature map effectively distills the complex non-linear temporal dynamics of the time series, as demonstrated by our empirical analysis. Additionally, we refine our model's hyperparameters using Bayesian optimization (BO) to circumvent the tedious gradient descent optimization, generally realized following the parameter-shift rule. To the best of our knowledge, this work represents the first application of GPR based on a quantum kernel to address the challenge of probabilistic TSF.

\textbf{Paper Outline.} The remainder of this paper is organized as follows. Section \ref{sec:background} introduces relevant background concepts. Section \ref{sec:Methodology} presents our modeling approach. Section \ref{sec:Experiments} outlines the empirical evaluation setup and our results. Finally, Section \ref{sec:Conclusion} concludes the paper.

\section{Background} \label{sec:background}
\subsection{Gaussian process regression}\label{subsec:GPR}
Consider a set of observations $\mathcal{D} = {(\mathbf{x}_i, y_i)}_{1 \leq i \leq c}$, with each $\mathbf{x} \in \mathbb{R}^w$ representing a $w$-dimensional input vector of covariates and $y \in \mathbb{R}$ signifying the scalar output or target. We organize these observations into a $w \times c$ design matrix $X$, and stack the corresponding target values into the vector $\mathbf{y}$, thereby representing the dataset as $\mathcal{D} = (X,\mathbf{y})$. In regression analysis, the goal is to model the input-target relationship through the function $y = f(\mathbf{x}) + \epsilon$, where $f: \mathbb{R}^{w} \mapsto \mathbb{R}$ is designed to approximate the unknown relationship, and $\epsilon$ is a normally distributed error term with zero mean and variance $\sigma_n^2$, denoted as $\epsilon \sim \mathcal{N}(0, \sigma_n^2)$.

Parametric models, such as linear regression, posit a specific functional form for $f$, typically involving a set of parameters $\mathbf{\mu}$, e.g., $f(\mathbf{x}) = \mathbf{x}^\intercal \mathbf{\mu}$, $\mathbf{\mu} \in \mathbb{R}^w$, for linear models. Conversely, GPR adopts a non-parametric approach, wherein $f$ is not constrained by a predefined structure but is instead described by a distribution over possible functions. This probabilistic treatment of the model function $f$ allows GPR to flexibly capture complex patterns in the data without committing to a fixed form, a significant advantage when our prior knowledge is limited or when the true underlying relationships are intricate and not well-understood.

To manage the intractable set of all possible functions from $\mathbb{R}^{w} \mapsto \mathbb{R}$, we impose a structure on the prior distribution of $f$, encapsulating it within the framework of a Gaussian process (GP). This GP, represented by the collection of random variables $\big( f(\mathbf{x}_1), f(\mathbf{x}_2), ..., f(\mathbf{x}_c) \big)$, is fully characterized by its mean function $m(\mathbf{x})$ and covariance function $\kappa(\mathbf{x}, \mathbf{x}')$. The mean function captures the expected value of the process, while the covariance function quantifies the extent to which the outputs at different inputs co-vary. Formally, we express these functions as $m(\mathbf{x}) = \mathbb{E}\big[f(\mathbf{x})\big]$ and $\kappa(\mathbf{x}, \mathbf{x}') = \text{Cov}\big(f(\mathbf{x}), f(\mathbf{x}') \big)$, denoting the process as $f(x) \sim \mathcal{GP}\big(m(\mathbf{x}), \kappa(\mathbf{x}, \mathbf{x}') \big)$.

The selection of the kernel function $\kappa$ profoundly influences the characteristics of the function class deemed likely to model the data. For instance, the radial basis function (RBF) kernel, expressed in \eqref{eq:rbf-kernel}, predisposes the model to smooth functions, while a periodic kernel, detailed in \eqref{eq:periodic-kernel}, biases it towards periodic functions. This critical choice of kernel is thus instrumental in steering the regression outcomes and warrants careful consideration to align with the nature of the data and the problem at hand.

\subsection{Quantum machine learning}\label{sec:QML}
QML is a burgeoning paradigm in artificial intelligence, leveraging the unique capabilities of quantum computers to handle computations within exponentially large vector spaces through quantum algorithms \cite{cerezo2021variational, schuld2021machine, zeguendry2023quantum}. In typical QML applications, classical datasets residing in some space $\mathcal{X}$ are used for tasks like classification or data generation. These datasets are transferred to the quantum domain using a feature map $\phi: \mathcal{X} \mapsto \mathcal{H}$, where $\mathcal{H}$ denotes the Hilbert space of an $n$-qubit quantum computer, with dimensions $2^n$. For each data point $\mathbf{x} \in \mathcal{X}$, embedding unitaries $\mathcal{U}(\mathbf{x})$ are applied such that $\mathcal{U}(\mathbf{x})\ket{0}^{\otimes n} = \ket{\phi(\mathbf{x})}$ produces a state dependent on the data. The model typically evolves by training a measurement operator $\mathcal{M}_\theta$ to minimize a loss function $\mathcal{L}$ over the dataset, with the model output defined as $h_\theta(\mathbf{x}) = \bra{\phi(\mathbf{x})} \mathcal{M}_\theta \ket{\phi(\mathbf{x})}$ \cite{schuld2021machine}.

In quantum kernel applications, one does without the trainable measurement circuit $\mathcal{M}_\theta$, and the quantum computer is instead used to compute similarity measures between pairs of data points $(\mathbf{x}, \mathbf{x}')$. Any efficient embedding unitary $\mathcal{U}(\mathbf{x})$ implicitly defines a fidelity kernel
\begin{equation}
    \kappa(\mathbf{x},\mathbf{x'}) = \bra{0}^{\otimes n} \mathcal{U}^{\dag}(\mathbf{x}')\mathcal{U}(\mathbf{x})\ket{0}^{\otimes n},
\end{equation}
which can be computed efficiently on a quantum computer by reversing the embedding unitary $\mathcal{U}(\mathbf{x'})$. For a large class of unitaries $\mathcal{U}$, simulating this kernel classically requires exponential resources, and is thus unfeasible for a relatively large number of qubits. Having access to a quantum computer therefore enables new classes of kernel functions, potentially improving the performance of kernel-based algorithms. In this study, we harness a quantum kernel function, conjectured to be classically hard to simulate, for GPR in time series forecasting.

\section{Methodology} \label{sec:Methodology}
\subsection{Problem statement} \label{subsec:Problem_statment}
Consider a time series $\mathbf{s}$, observed over the time interval $\mathbb{T} = \{t_l: l \in [1 .. T]\}$, where each $t_l$ is a positive real number and $T$ is a positive integer. This series comprises a sequence of real-valued observations $x_t$, systematically recorded in chronological order as expressed by:
\begin{equation}\label{eq:timeSeriesDefinition}
    \mathbf{s} = (x_t : t \in \mathbb{T}), \;\;\; x_t \in \mathbb{R}.
\end{equation}

\textbf{Problem.} Our focus is on the task of probabilistic forecasting, aiming to predict the forthcoming value $x_{t_{l+1}}$ in the series $\mathbf{s}$, based on a historical window of observations of length $w$: $(x_{t}: t_{l-w+1}\leq t \leq t_{l})$. The challenge lies in accurately determining the conditional probability distribution of the next observation as follows:
\begin{equation} \label{eq:condProbDist}
   p\bigg( x_{t_{l+1}} \;|\; (x_{t}: t_{l-w+1}\leq t \leq t_{l})\bigg),
\end{equation}
ensuring the continuity of the sequence where the lookback period starts after $t_1$ and the prediction point $t_{l+1}$ falls within the observation period ending at $t_T$.

\subsection{Forecasting as a regression problem} \label{subsec:forecasting-regression}
We conceptualize our forecasting challenge within the regression framework as follows:
\begin{equation}
    y = f(\mathbf{x}) + \epsilon,
\end{equation}
where $y$, the target value, is defined as $y = x_{t_{l+1}}$ and represents the next value in the time series. The input vector $\mathbf{x} \in \mathbb{R}^{w}$ comprises a sequence of $w$ preceding observations $(x_{t}: t_{l-w+1}\leq t \leq t_{l})^\intercal$, and $\epsilon$ denotes an independent and identically distributed normal additive noise, specifically $\epsilon \sim \mathcal{N}(0, \sigma_n^2)$. Drawing from the GPR framework \cite{rasmussen2006gaussian}, we postulate that any set of input vectors $\mathcal{X} = \{\mathbf{x}_1, \mathbf{x}_2,...,\mathbf{x}_c\}$, extracted from the time series $\mathbf{s}$ and numbering $c$ in total, induces a stochastic process $\big(f(\mathbf{x}_{1}), f(\mathbf{x}_{2}),...,f(\mathbf{x}_{c}) \big)$ that conforms to a Gaussian distribution. Here, each $f(\mathbf{x}_j)$ is a random variable, with randomness originating from the selection of the function $f$ from a space of possible functions $\mathbb{R}^{w} \mapsto \mathbb{R}$.

This Gaussian process is delineated by its prior distribution, defined by a mean function $m(\mathbf{x})$—typically assumed constant (i.e., $m(\mathbf{x}) = m$ with $m$ as a hyperparameter, usually set to zero)—and a covariance kernel function $\kappa_{\alpha}(\mathbf{x}_{j}, \mathbf{x}_{j'})$ reliant on some hyperparameters $\alpha \in \mathbb{R}^{d_{\alpha}}$, where $d_{\alpha} \in \mathbb{Z}^+$ is the number of such hyperparameters. Thus, the prior distribution for a given set of input windows $\mathcal{X}$ is expressed as
\begin{equation}\label{GaussianProcess}
    f(\mathbf{x}) \sim \mathcal{GP}\big(m, \kappa_{\alpha}(\mathbf{x}_{j}, \mathbf{x}_{j'})\big).
\end{equation}

Throughout this discussion, we employ $f(\mathbf{x})$ and $f$ interchangeably to refer to the model's output, thereby treating it as a random variable influenced by the underlying probabilistic nature of the GPR model.

\subsection{Predictive distribution}\label{subsec:predictiveDistribution}
In the context of a new observation window $\mathbf{x}$, and considering a set of training windows $\mathcal{X} = \{\mathbf{x}_{1}, \mathbf{x}_{2},..., \mathbf{x}_{c} \}$ with their corresponding targets $\mathcal{Y} = \{y_{1}, y_{2},..., y_{c} \}$, the posterior distribution of the forecasted values $f$, conditioned on both the training windows and the new input sequence $\mathbf{x}$, is captured by the predictive distribution
\begin{equation}\label{PredictiveDist}
    f \;|\; X, \mathbf{y}, \mathbf{x}  \sim \mathcal{N}\big(\bar{f}, \sigma_f^2\big),  
\end{equation}
where $X$ denotes the matrix compiled from the vectors $\mathbf{x}_{j}$, arranged as columns such that $X = (\mathbf{x}_{1}, \mathbf{x}_{2},..., \mathbf{x}_{c})$ and occupies a space in $\mathbb{R}^{w \times c}$. The vector $\mathbf{y}$ consolidates the target values into $(y_{1}, y_{2},..., y_{c})^\intercal$ within $\mathbb{R}^{c}$. Here, $\bar{f}$ represents the mean prediction, and $\sigma_f^2$ denotes the forecast's variance. Such distribution parameters are derived in \cite{rasmussen2006gaussian} as:
\begin{equation}\label{PredictiveDistribution-parameters}
    \begin{split}
        \bar{f} &= \mathbf{k}^\intercal \left(K + \sigma_n^2 I \right)^{-1} \mathbf{y} + m, \\
        \sigma_f^2 &= \kappa_{\alpha}(\mathbf{x}, \mathbf{x}) - \mathbf{k}^\intercal \left( K + \sigma_n^2 I \right)^{-1} \mathbf{k},    
    \end{split}
\end{equation}
where $\mathbf{k}$ is a vector in $\mathbb{R}^c$, formed by the covariance values between the new observation window $\mathbf{x}$ and each of the training windows $\mathbf{x}_{j}$, specifically $\mathbf{k} = (\kappa_{\alpha}(\mathbf{x}, \mathbf{x}_{j}): j \in [1..c] )^\intercal$. The covariance matrix $K$, a square matrix of dimensions $c \times c$, is described as $K = K(X,X) = (\kappa_{\alpha}(\mathbf{x}_{i}, \mathbf{x}_{j}))_{i,j = 1}^{c}$, encapsulating the pairwise covariances between all training window combinations. This formulation elegantly encapsulates the uncertainty and expectations of model predictions, providing a comprehensive probabilistic outlook on forecasting. 

\subsection{Kernel function}
Our method employs a strategy of constructing the kernel function in a bottom-up manner using sub-windows. This approach enables the generation of embeddings that capture the intricate temporal patterns within each subsequence—a nuance that methods focusing on individual time point embeddings might miss \cite{baker2023massively}.

\subsubsection{Quantum encoding feature map} \label{subsubsection:quantum-encoding-feature-map}
\begin{figure}
    \centering
    \includegraphics[width=1\linewidth, height=0.2\textheight]{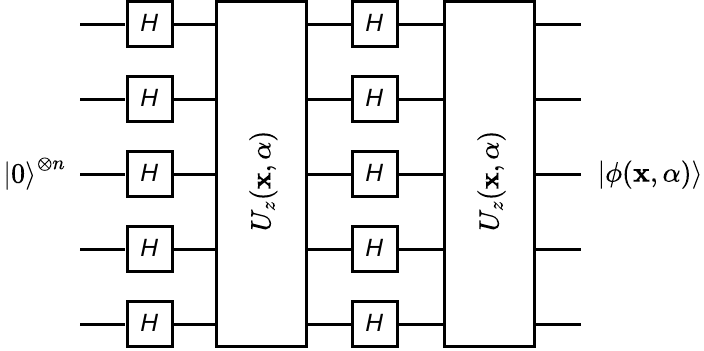}
    \caption{The IQP feature map $\mathcal{U}(\mathbf{x}, \alpha)$ constructs the quantum state $\ket{\phi(\mathbf{x}, \alpha)}$ by sequentially applying two layers of operations across all qubits: first, Hadamard gates are applied simultaneously to each qubit, followed by the collective application of the unitary $U_z(\mathbf{x}, \alpha)$.}
    \label{fig:iqp_feature_mapl}
\end{figure}

Consider a time series window $\mathbf{x} = (x_{t_1}, x_{t_2}, ..., x_{t_w}), t_i \in \mathbb{T}$. We map this sequence into a quantum state via the data encoding feature map $\phi$, producing a quantum state $\ket{\phi(\mathbf{x}, \alpha)}$ as follows:
\begin{equation}
    \phi : \begin{aligned} 
    \mathbb{R}^{w} &\mapsto \mathcal{H} \\
    \mathbf{x} & \mapsto \ket{\phi(\mathbf{x}, \alpha)},
    \end{aligned}
\end{equation}
where $\alpha \in \mathbb{R}$, is introduced as a tunable hyperparameter. 
The state $\ket{\phi(\mathbf{x}, \alpha)}$ is prepared by applying the embedding unitary $\mathcal{U}(\mathbf{x}, \alpha)$ to the initial state $\ket{0}^{\otimes n}$, as shown by the equation 
\begin{equation}
    \mathcal{U}(\mathbf{x}, \alpha) \ket{0}^{\otimes n} = \ket{\phi(\mathbf{x}, \alpha)}.
    \label{eq:feature-map}
\end{equation}

\subsubsection{IQP feature map}\label{subsubsec:IQP-feature-map}

For our feature map to take full
advantage of the capacity of quantum computers, it must leverage quantum properties challenging for classical computation to simulate.
Accordingly, we have designed our embedding strategy to incorporate an Instantaneous Quantum Polynomial-time (IQP)-style circuit, echoing the proposals for achieving quantum advantage introduced in \cite{havlivcek2019supervised}.

The architecture of our feature map $\mathcal{U}(\mathbf{x}, \alpha)$, depicted in Fig.~\ref{fig:iqp_feature_mapl}, is described by the equation:
\begin{equation}\label{eq:iqp-feature-map}
    \ket{\phi(\mathbf{x}, \alpha)} = U_z(\mathbf{x}, \alpha) H^{\otimes n} U_z(\mathbf{x}, \alpha) H^{\otimes n} \ket{0}^{\otimes n},
\end{equation}
where $H^{\otimes n}$ denotes the Hadamard gates applied simultaneously to all qubits. The unitary $U_z(\mathbf{x}, \alpha)$ is defined as: 
\begin{equation}\label{eq:iqp-unitary}
    U_z(\mathbf{x}, \alpha) = \text{exp}\left(\alpha \sum_{j=1}^{n} x_{t_j} Z_j + \alpha^2 \sum_{j'< j=1}^{n} x_{t_j} x_{t_{j'}} Z_j Z_{j'} \right),
\end{equation}
with $Z_j$ being the Pauli-Z gate affecting the j-th qubit, and $\alpha \in [0,1]$ representing the bandwidth coefficient. This coefficient is introduced to improve the model's generalization capability by confining the embeddings within a limited region of the quantum state space, thereby mitigating the risk of generalization errors in expansive Hilbert spaces \cite{canatar2022bandwidth, shaydulin2022importance}. It is assumed that the input sequence $\mathbf{x}$ will be normalized during data preprocessing to center around zero with a standard deviation of one, aligning with the initial proposal's framework \cite{havlivcek2019supervised}.

This quantum feature map aims to nonlinearly unveil the intricate temporal dynamics inherent in the sequence $\mathbf{x}$. By translating classical data into quantum information, the feature map captures not only individual temporal contributions but also their interactive dynamics. These interactions, resembling Ising-type interactions, encode temporal correlations across different time points, enabling a sophisticated understanding of the temporal structure in $\mathbf{x}$.

The ability of this embedding to entangle subsystems contributes significantly to its classical intractability. Entanglement arises through the application of the $Z_j Z_{j'}$ unitaries, inducing correlations between qubit pairs and thus intertwining different subsystems. Additionally, the circuit's dual-layer architecture plays a crucial role in its complexity, as demonstrated in \cite{havlivcek2019supervised}, since a single-layer circuit may still permit classical estimation, underscoring the nuanced design of our feature map for harnessing quantum advantage.

\subsubsection{The kernel function}
Employing the embedding feature map $\mathbf{x} \mapsto \ket{\phi(\mathbf{x}, \alpha)}$, we quantify the similarity between sequences $\mathbf{x}$ and $\mathbf{x}'$ by calculating the fidelity between their quantum embeddings, $\ket{\phi(\mathbf{x}, \alpha)}$ and $\ket{\phi(\mathbf{x}', \alpha)}$. This measure, rooted in established literature \cite{schuld2021supervised, havlivcek2019supervised}, is defined as:
\begin{equation}\label{eq:raw-fidelity-kernel}
\kappa_{\alpha}(\mathbf{x}, \mathbf{x}') = |\braket{\phi(\mathbf{x}, \alpha) | \phi(\mathbf{x}', \alpha)}|^2.
\end{equation}
Crucially, this kernel function fulfills the symmetry requirement, a core attribute for kernel validity, guaranteeing that $\kappa_{\alpha}(\mathbf{x}, \mathbf{x}') = \kappa_{\alpha}(\mathbf{x}', \mathbf{x})$ holds for any pair of sequences $\mathbf{x}$ and $\mathbf{x}'$.

\begin{figure}
    \centering
    \includegraphics[width=0.8\linewidth, height=0.2\textheight]{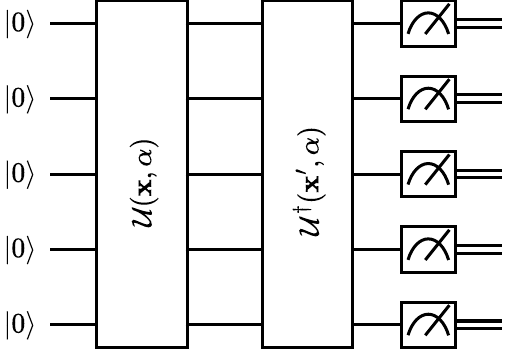}
    \caption{Quantum kernel realization via fidelity state overlap, wherein the circuit evaluates fidelity by sequentially applying unitary $\mathcal{U}(\mathbf{x}, \alpha)$, its inverse $\mathcal{U}^\dag(\mathbf{x}', \alpha)$, and estimating the probability of an all-zero measurement outcome.}
    \label{fig:realization_fildeity}
\end{figure}

Fig.~\ref{fig:realization_fildeity} illustrates the implementation of such a quantum kernel. The circuit computes the fidelity between the quantum states in the latent space by first applying the unitary transformation $\mathcal{U}(\mathbf{x}, \alpha)$ and then its inverse $\mathcal{U}^\dag(\mathbf{x}', \alpha)$. This is followed by measuring all qubits at the circuit's conclusion \cite{havlivcek2019supervised}. The probability of observing zero across all output bits, denoted by $p_{0, ..., 0}$, directly corresponds to the overlap of the desired states, hence $p_{0, ..., 0} = |\braket{\phi(\mathbf{x}, \alpha)| \phi(\mathbf{x}', \alpha)}|^2$. 

\subsection{Hyperparameters fine-tuning} \label{subsec:optimization}
Optimizing hyperparameters $\theta \in \mathbb{R}^3$, which encompass the Gaussian Process (GP)'s mean constant $m$, noise variance $\sigma_n^2$, and bandwidth $\alpha$, is paramount in our model. The objective function $g(\theta)$, designated for optimization, is the marginal log likelihood (MLL). This function quantifies the model's fit by marginalizing over the set of all possible functions $\mathbf{f} \in \mathbb{R}^c$:
\begin{equation} \label{eq:mll}
    \log p_{\theta}(\mathbf{y} | X) = \log \int_{\mathbf{f}} p_{\theta}(\mathbf{y} | \mathbf{f}, X) p_{\theta}(\mathbf{f} | X).
\end{equation}
Within the Gaussian process framework, with the prior on $\mathbf{f}$ being Gaussian, $\mathbf{f} | X \sim \mathcal{N}(m, K)$, and the likelihood as a factorized Gaussian, $\mathbf{y} | \mathbf{f}, X = \mathbf{y} | \mathbf{f} \sim \mathcal{N}(\mathbf{f}, \sigma_n^2
I)$, this integral assumes a tractable form \cite{rasmussen2006gaussian}.

While gradient descent could be a natural choice for its precision, the computational demands, particularly the increased number of quantum circuit runs required by the parameter shift rule \cite{mitarai2018quantum}, make it less appealing. Additionally, a comprehensive exploration of the parameter space using gradient descent demands a large number of evaluations. Thus, we adopt Bayesian optimization for its efficiency with expensive objective functions and its derivative-free nature. Further details about the Bayesian optimization framework can be found in appendix \ref{app:BO}.

For this optimization process, we employ another GP model—distinct from the one used for TSF—to serve as our BO's surrogate model, selecting the Matérn 5/2 kernel, expressed in \eqref{eq:matern-kernel}, for its flexibility and capability to model less smooth surfaces compared to the RBF kernel \cite{rasmussen2006gaussian}. The log expected improvement is chosen as the acquisition function for its numerical optimization ease \cite{ament2024unexpected}.

The initial phase of our BO utilizes a Sobol sequence for space-filling design, generating quasi-random sequences that more uniformly cover the hyperparameter space than pure random sampling. In the subsequent phase, we enter a loop where, at each step $1 \leq j \leq N$, a new query point $\theta_j$ is selected to optimize the acquisition function, i.e.,
\begin{equation} \label{eq:inner-optimz-pb}
    \theta_j = \underset{\theta \in \Theta}{\arg \max} \; \text{EI}_{z_j^*}(\theta),
\end{equation}
with $z_j^*$ representing the highest observed value so far, and $\mathcal{D}_j$ the dataset of observations made so far.

The optimization bounds are specifically set for each parameter: $0 \leq \alpha \leq 1$ for the bandwidth, $m_0 \leq m \leq m_1$ for the mean constant, and $\sigma_{n_0}^2 \leq \sigma_n^2 \leq \sigma_{n_1}^2$ for the noise level, where the bounds are user-defined. The problem defined in \eqref{eq:inner-optimz-pb} is tackled through an inner optimization using the Limited-memory Broyden–Fletcher–Goldfarb–Shanno with Bounds (L-BFGS-B) algorithm, preferred for bound-constrained optimization \cite{byrd1995limited, zhu1997algorithm}. Upon determining the next query point $\theta_j$, the function $g$ is evaluated, added to the set of observations, and the surrogate GP model is updated, paving the way for precise and efficient hyperparameter tuning.

\section{Experiments} \label{sec:Experiments}

\begin{figure} 
    \centering
    \includegraphics[width=0.8\linewidth]{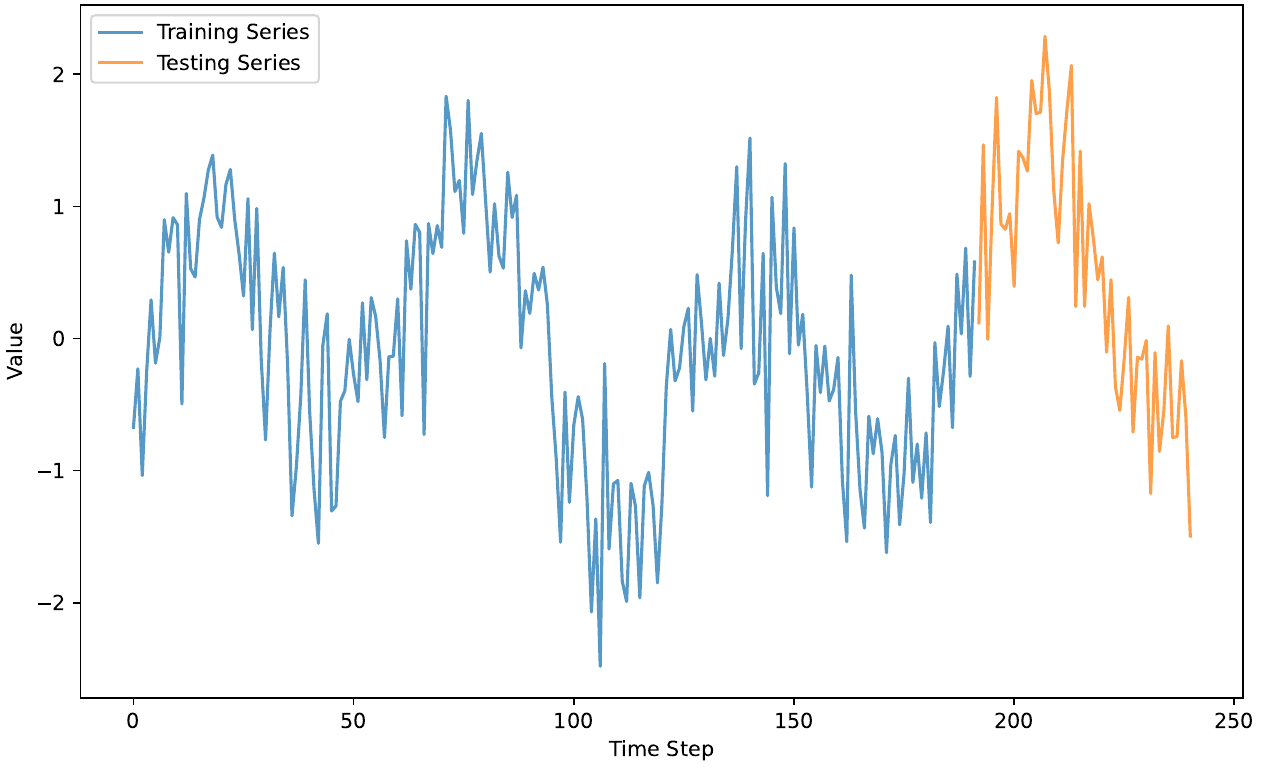}
    \caption{The synthetic time series, featuring multiple non-linear patterns, is divided into two sets: the training set $\mathcal{D}$, shown as the blue curve, and the testing set $\mathcal{D}'$, represented by the orange curve.}
    \label{fig:generated_synthetic_series}
\end{figure}

To test our quantum kernel-based GPR model, we have opted for a synthetic time series to build intuition. Our goal is to create a synthetic TSF problem where (i) data is not linearly forecasted (thus kernelized approaches are required), (ii) the dynamics of the time series are not static and evolve with time in order to mimic real world time series. To this end, we have generated a synthetic time series, illustrated in Fig.~\ref{fig:generated_synthetic_series}, that combines linear trend components, non-linear dynamics through sine waves, and Gaussian noise to create complex patterns for testing and analysis. Specifically, for a given number of time steps (240 our case), it first divides the series into segments according to the specified number of trend changes (4), alternating between upward and downward linear trends with slight slopes. In parallel, it overlays two sine waves: one with a longer period (10 time steps) and larger amplitude (scale of 1) to introduce long-term cyclic patterns, and another with a period determined by the frequency of trend changes and a smaller amplitude (scale of 0.5) to capture quicker, more nuanced oscillations. The resultant series is further perturbed by adding Gaussian noise with a specified level (0.5), to mimic real-world unpredictability. This process results in a synthetic dataset that exhibits realistic characteristics such as trends, seasonality, and noise, making it useful for testing TSF models.

Adhering to the preprocessing steps outlined in \cite{huang2021mcclean}, we commence our experimental procedure by standardizing the generated synthetic time series. This normalization process adjusts the series values to be centered around zero with a standard deviation of one. Subsequently, the series is segmented into training and testing sets, depicted by the blue and orange curves, respectively, in Fig.~\ref{fig:generated_synthetic_series}. The training dataset is denoted as $\mathcal{D} = (\mathcal{X}, \mathcal{Y})$, where $\mathcal{X} = \{\mathbf{x}_1, ..., \mathbf{x}_c\}$ encompasses the input windows, and $\mathcal{Y} = \{y_1, ..., y_c\}$ contains the corresponding target values. In our experiments, we selected a window length of 5, necessitating the use of 5 qubits per window, aiming for a balanced qubit count that strikes a compromise between being overly extensive and unduly limited.
To minimize redundancy in the training data, consecutive training windows $\mathbf{x}_i$ and $\mathbf{x}_{i+1}$ are allowed an overlap of merely 2 time steps. The test dataset, $\mathcal{D}' = (\mathcal{X}', \mathcal{Y}')$, is structured to commence prediction at the immediate subsequent time point following the last prediction in the training phase. The prediction process then advances by a single time step, sequentially addressing the remaining time points in the series until its conclusion. 

\subsection{Results of the quantum kernel}\label{subsec:results-qk}

In our exploration, the quantum-based GPR model, trained on a classical simulator, was fine-tuned to fit 
the synthetically generated time series data employing the BO technique as detailed previously. The hyperparameters optimization was structured into two key steps: an initialization phase and a query point generation phase, each with 25 sample points, i.e., $n_0 = N = 25$, to enhance the model's predictive accuracy. The hyperparameter selection was meticulously designed to maximize the marginal log likelihood, ensuring the model's optimal performance during test set predictions.

\begin{figure} 
    \centering
    \includegraphics[width=0.8\linewidth]{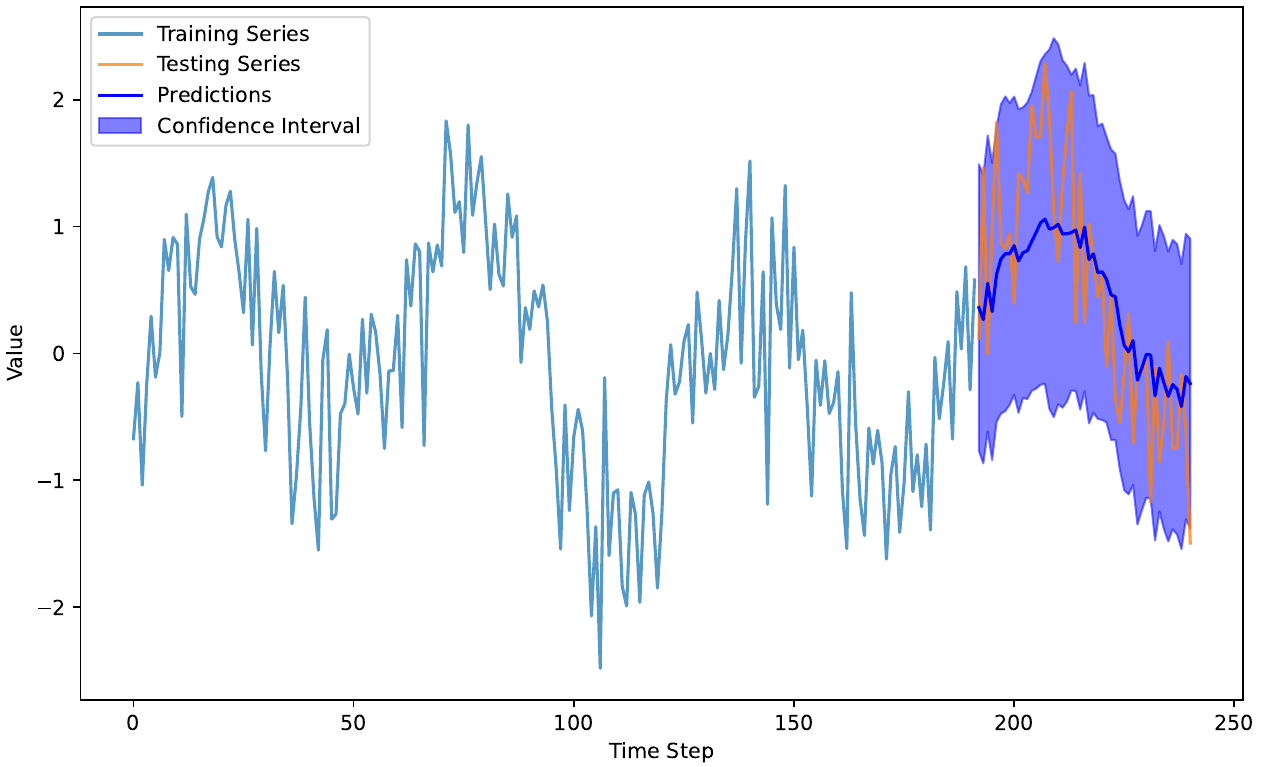}
    \caption{The IQP-based GP's mean predictions are depicted by the dark blue curve, with the light shaded area indicating the predictions' 95\% confidence interval.}
    \label{fig:test_predictions_iqp}
\end{figure}

Fig.~\ref{fig:test_predictions_iqp} showcases the predictive prowess of our model on the test set, where the training data is illustrated by a light blue curve, the test data by an orange curve, and the model’s predictions by a dark blue curve, surrounded by a lightly shaded area indicating the 95\% confidence interval. Observably, the model exhibits a good fit to the test series, with the confidence interval accommodating the series' variability. This illustrates the quantum kernel's aptitude in capturing the nonlinear dynamics of the series while maintaining flexibility, highlighting its effectiveness in forecasting intricate time series data.

Throughout the optimization journey, the objective function, known for its computational expense, was evaluated 50 times to fine-tune the hyperparameters. The optimization bounds for the GPR model's mean constant $m$ and the noise level $\sigma_n^2$ were set to $-1 \leq m \leq 1$ and $0 \leq \sigma_n^2 \leq 1$, respectively. This decision was influenced by the noise level in the synthetic dataset, set at 0.5, and the mean constant $m = 0$, derived from the data normalization process, hence selecting intervals that encompass these values. 

Ultimately, the optimal configuration for our quantum-based GPR model was identified as $\theta = (\alpha, \sigma_n^2, m)^\intercal = (0.243, 0.350, 0.503)^\intercal$. 
The optimal bandwidth parameter $\alpha = 0.243$ suggests a strategic limitation of the latent Hilbert space, signifying that our kernel could effectively model the time series data with restrained expressibility. This constraint acts as a form of regularization, aiding in the learning process and preventing model overfitting.

\begin{figure}
    \centering
    \subfloat[Contour plot $(\alpha, m)$.\label{fig:contour-plot-alpha-m}]{
        \includegraphics[width=0.4725\linewidth]{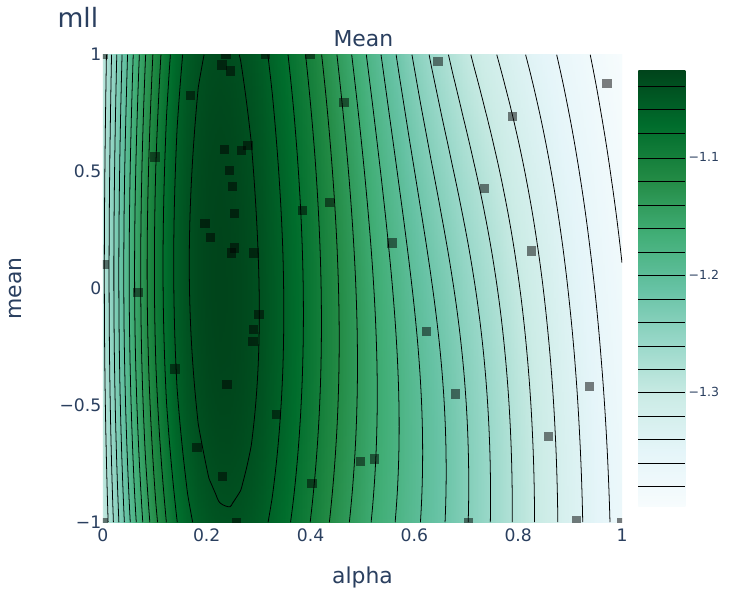}
    }
    \subfloat[Contour plot $(\alpha, \sigma_n^2$).\label{fig:contour-plot-alpha-sigma}]{
        \includegraphics[width=0.4725\linewidth]{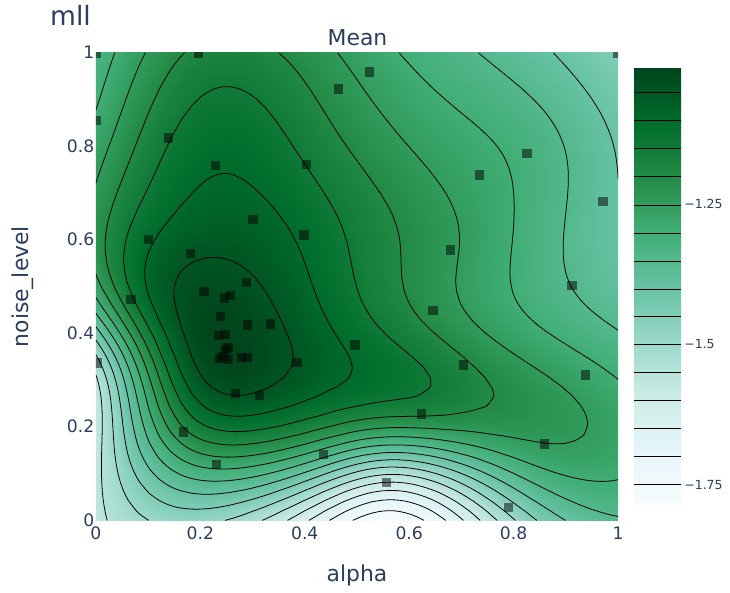}
    }
    \caption{Bayesian optimization contour plots show evaluated configurations from two perspectives with a color gradient from white to dark green representing MLL values—darker shades indicate higher values, and square markers denote objective function evaluations.}
    \label{fig:iqp_bo_process}
\end{figure}

Fig.~\ref{fig:iqp_bo_process} 
visualizes the various configurations, evaluated during the optimization process, viewed in two ways, with a color gradient from white to dark green denoting the marginal log likelihood values—darker shades represent higher values, and square markers indicate the evaluation of the objective function. The sub-figures are two views of the same process plotting the MLL against different sub-configurations of hyperparameters $(\alpha, m)$ and $(\alpha, \sigma_n^2)$. Post-initialization, which employed a Sobol sequence for diverse hyperparameter sampling, the model progressively focused on configurations within the higher-value (dark green) domain, reflecting a sophisticated convergence towards the most promising hyperparameter region.

\subsection{Quantum kernel vs baseline classical kernels} \label{subsec:qkernel-vs-classicals}

\subsubsection{Baseline kernels}

To evaluate the effectiveness of our proposed quantum kernel, we conducted comparisons against a selection of widely recognized classical kernels in the literature. To this end, we have selected four classical baseline kernels for comparison, providing a brief overview of each below.

Our first comparison involves the Radial Basis Function (RBF) kernel, which is notably prevalent in classical machine learning literature \cite{vert2004primer}. It quantifies the similarity between two samples, $\mathbf{x}, \mathbf{x}' \in \mathbb{R}^w$, through the equation
\begin{equation} \label{eq:rbf-kernel}
    \kappa_{\text{RBF}_{l_r}}(\mathbf{x}, \mathbf{x}') = \exp \left(- \frac{\lVert \mathbf{x} - \mathbf{x}' \rVert_{\ell_2}^2 }{2 l_r^2} \right),
\end{equation}
where $l_r$ denotes the lengthscale hyperparameter, constrained between $0.1$ and $30$ in our experiments, and $\lVert \mathbf{x} - \mathbf{x}' \rVert_{\ell_2}^2$ represents the Euclidean distance between the two feature vectors.

Next, we consider the Matérn kernel \cite{genton2001classes}, which models the covariance between $\mathbf{x}, \mathbf{x}' \in \mathbb{R}^w$ as
\begin{equation} \label{eq:matern-kernel}
    \kappa_{\text{MAT}_{\nu, l_m}} = \frac{2^{1-\nu}}{\Gamma(\nu)} \left( \sqrt{2\nu} d \right)^\nu K_\nu \left( \sqrt{2\nu} d \right),
\end{equation}
where $d$ is the scaled Euclidean distance between $\mathbf{x}$ and $\mathbf{x}'$, $l_m$ is the lengthscale hyperparameter (with a constraint of $0.1 \leq l_m \leq 30$), $\nu$ indicates the smoothness parameter with values in ${1/2, 3/2, 5/2}$—affecting the kernel's smoothness, and $K_\nu$ is the modified Bessel function.

\begin{figure*}
    \centering
    \subfloat[IQP predictions.\label{subfig:IQP}]{
        \includegraphics[width=0.32\textwidth]{figs/test_predictions_plots/test_predictions_iqp.pdf}
    }
    \subfloat[RBF predictions.\label{subfig:RBF}]{
        \includegraphics[width=0.32\textwidth]{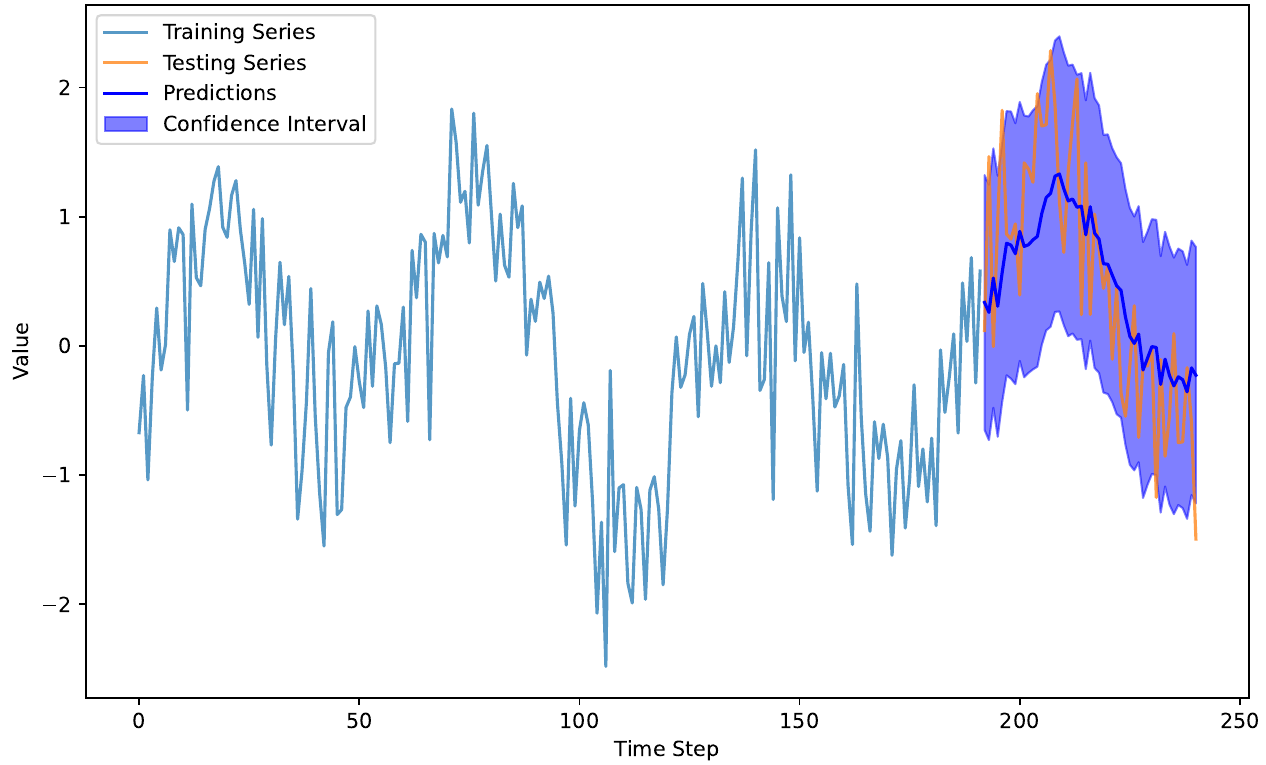}
    }
    \subfloat[Matérn predictions.\label{subfig:Matern}]{
        \includegraphics[width=0.32\textwidth]{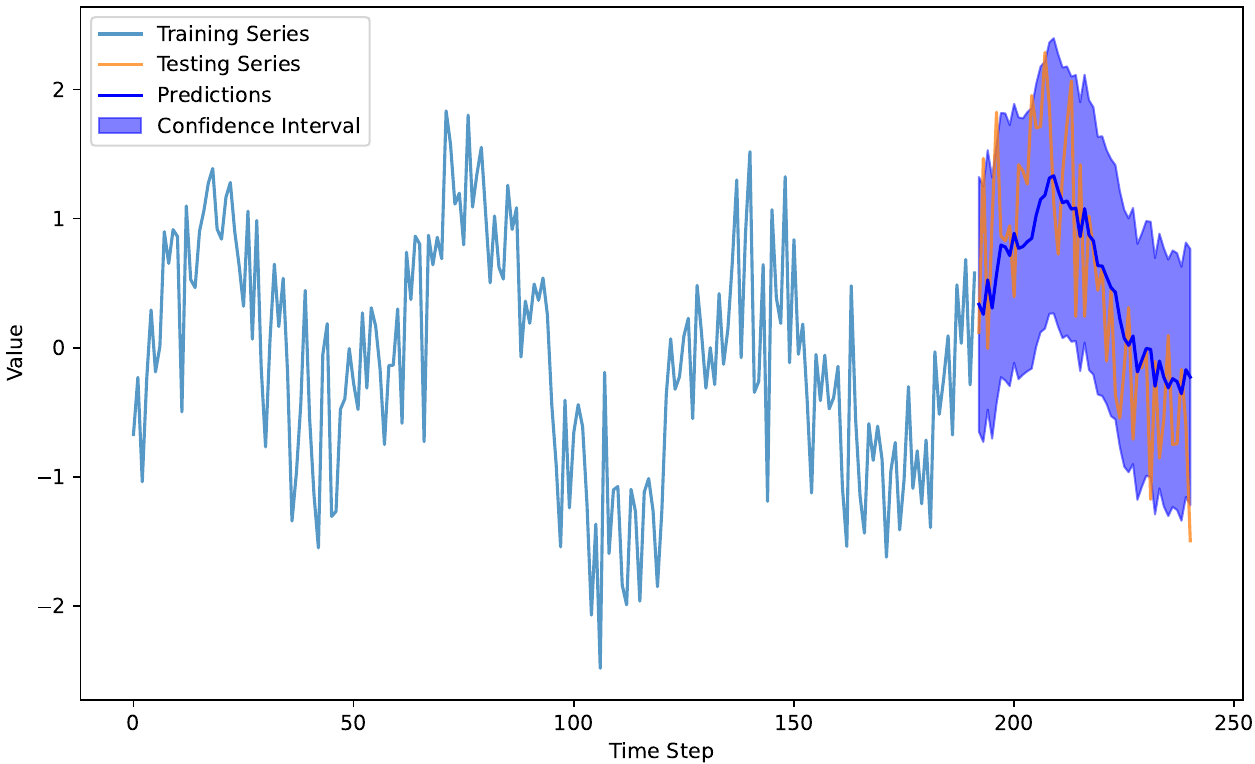}
    }
    
    \vspace{0.5em} 
    
    \subfloat[RQ predictions.\label{subfig:RQ}]{
        \includegraphics[width=0.32\textwidth]{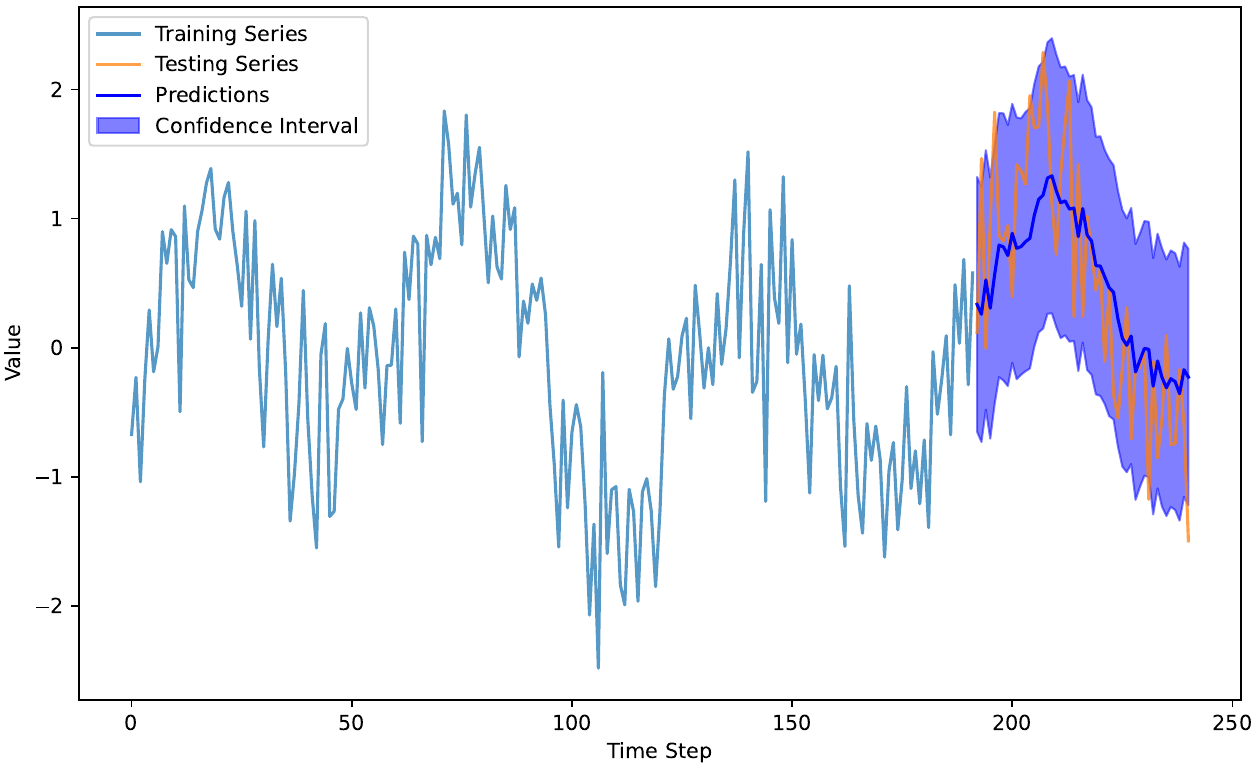}
    }
    \subfloat[Periodic predictions.\label{subfig:PER}]{
        \includegraphics[width=0.32\textwidth]{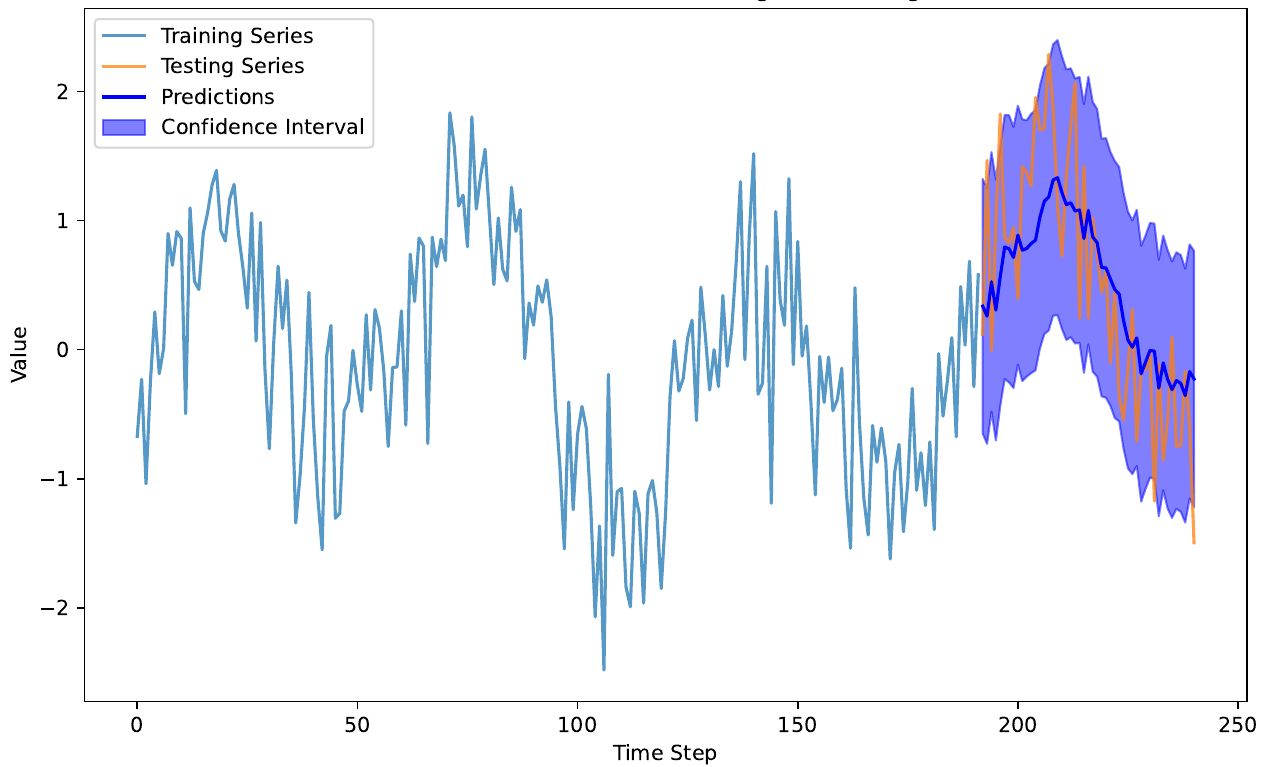}
    }
    
    \caption{Test prediction plots from various kernels show generally similar results, with the IQP model yielding a smoother mean predictions curve compared to other classical models.}
    \label{fig:test-predictions-different-kernels}
\end{figure*}

For the third kernel, we examine is the Rational Quadratic (RQ) kernel \cite{duvenaud2014kernel}, which measures similarity as
\begin{equation} \label{eq:rq-kernel}
    \kappa_{\text{RQ}_{\beta, l_q}}(\mathbf{x}, \mathbf{x}') = \left( 1 + \frac{\lVert \mathbf{x} - \mathbf{x}' \rVert_{\ell_2}^2 }{2 \beta l_q^2} \right)^{- \beta},
\end{equation}
where $\beta$ serves as the rational quadratic relative weighting parameter (ranging from $0.1$ to $10$), and $l_q$ is the lengthscale hyperparameter, constrained between $0.1$ and $30$.

Finally, our analysis includes the Periodic kernel \cite{mackay1998introduction}, represented by
\begin{equation} \label{eq:periodic-kernel}
    \kappa_{\text{PER}_{p, l_p}}(\mathbf{x}, \mathbf{x}') = \exp \left( -2 \sum_{i=1}^{w} \frac{\sin^2 \left( \frac{\pi}{p} (x_i - x'_i)\right)}{l_p} \right),
\end{equation}
where $p$ indicates the period length parameter (set between $5$ and $35$), and $l_p$ is the lengthscale hyperparameter, with a constraint range of $0.1$ to $30$.

It is worth noting that the constraints on the classical kernels' hyperparameters were carefully chosen to accommodate the diverse periodicities, characteristics, and fluctuations observed in the synthetic time series generated for our study.

\subsubsection{Quantum-classic kernels comparison results}

\begin{table*}
\centering
\caption{Performance metrics for each kernel model, with the best outcomes in bold and the second-best underlined. Lower values indicate better performance for all metrics except the LL, where higher is better.}
\label{tab:performance-comparison}
{\small
\begin{tabularx}{\textwidth}{@{}l *{6}{X} r r@{}}
\toprule
\textbf{Kernel name} & \textbf{sMAPE} & \textbf{WAPE} & \textbf{MAPE} & \textbf{RMSE} & \textbf{MAE} & \textbf{MSE} & \textbf{mCRPS} & \textbf{LL} \\
\midrule
\midrule
IQP    & \underline{0.667528} & \underline{0.610769} & 0.940599 & \underline{0.687006} & \underline{0.561709} & \underline{0.471977} & \underline{0.394675} & \textbf{-50.173489} \\
RBF    & \textbf{0.653616} & \textbf{0.598063} & 0.942324 & \textbf{0.672914} & \textbf{0.550024} & \textbf{0.452813} & \textbf{0.391894} & -52.599395 \\
Matérn & 0.685362 & 0.621422 & 0.950523 & 0.696314 & 0.571506 & 0.484853 & 0.401341 & \underline{-50.216731} \\
RQ     & 0.695145 & 0.623391 & \textbf{0.903838} & 0.703741 & 0.573317 & 0.495251 & 0.403775 & -50.442423 \\
Periodic & 0.688907 & 0.617823 & \underline{0.906199} & 0.698173 & 0.569024 & 0.487445 & 0.400667 & -50.392473 \\
\bottomrule
\end{tabularx}
}
\end{table*}

The predictive capabilities of both classical and quantum kernels, each fine-tuned using the BO process for hyperparameter tuning, are illustrated in Fig.~\ref{fig:test-predictions-different-kernels}. Notwithstanding minor variations in the shape of the confidence intervals, the quantum kernel's predictions broadly align with those generated by classical counterparts. This alignment suggests that our quantum model performs competitively with classical alternatives. Furthermore, our model produces a smoother and slightly different curve for mean predictions compared to classical counterparts, notably demonstrated by the reduced sharpness of the peak around time step 210. This suggests that our quantum feature map provides a different perspective on the data, potentially enriching the representation tools at our disposal.

To expand our initial findings, we employed a suite of performance metrics to assess the efficacy of each kernel based on predictions made on the test set $\mathcal{D}'$. The chosen metrics include the Mean Squared Error (MSE), the Root Mean Squared Error (RMSE), the Mean Absolute Percentage Error (MAPE), the symmetric MAPE (sMAPE), and the Weighted Average Percentage Error (WAPE). These metrics serve to quantify the discrepancy between the mean predictions and the actual target values, thereby evaluating the forecast quality. Furthermore, to gauge the model's proficiency in accurately representing the observed data, we incorporated two additional metrics: the mean Continuous Ranked Probability Score (mCRPS) and the Log Likelihood (LL). Detailed explanations of these metrics and their computational methodologies are available in appendix \ref{app:performance-metrics}.

The comparative results are summarized in Tab. \ref{tab:performance-comparison}. Notably, most metrics employed are loss functions where lower values indicate superior performance, except for the Log Likelihood (LL), where higher values denote better fit. 
The table illustrates that our quantum kernel holds its ground against classical benchmarks, outperforming all but the RBF kernel across nearly all metrics. Moreover, it achieves the highest LL metric score, indicating a slightly better explanation of the test data.

Drawing from both the visual comparison in the prediction plots and the quantitative analysis via performance metrics, it is evident that our model adeptly captures the nonlinear temporal patterns inherent in the data series. This proficiency likely originates from the qubit-to-qubit Ising interactions integral to our quantum feature map \eqref{eq:iqp-feature-map}, underlining the quantum model's robust capability in TSF.

\subsection{Ablation study}
\subsubsection{Fidelity state overlap analysis}
\begin{figure}
    \centering
    \includegraphics[width=0.7\linewidth, height=0.2\textheight]{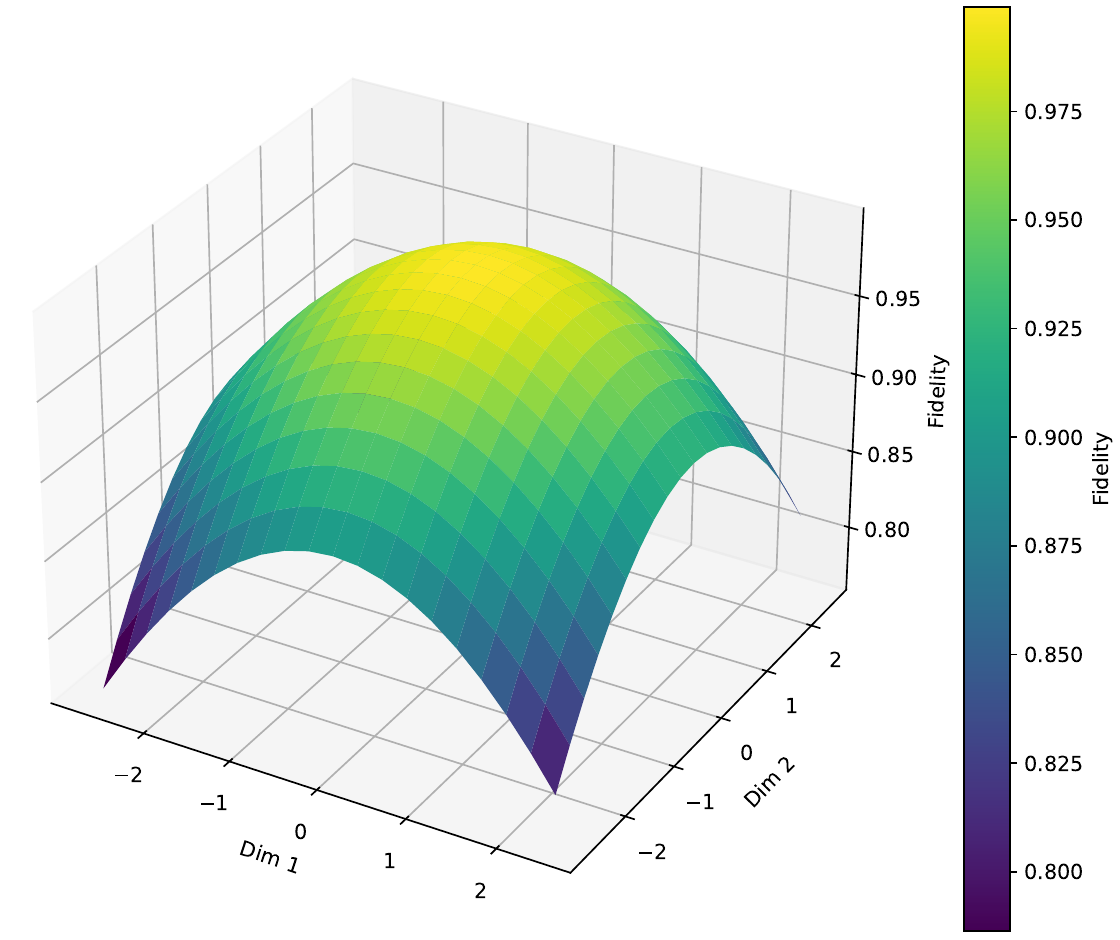}
    \caption{A three-dimensional plot depicting the fidelity state overlap between a centrally-located reference point and various sample points, demonstrating the efficacy of our quantum kernel. Closer proximity to the center correlates with higher fidelity, affirming the kernel's robustness.}
    \label{fig:state-overlap}
\end{figure}

To evaluate the effectiveness of our learned feature map, we delve into the fidelity state overlap among various quantum states, utilizing the kernel values as a proxy for this overlap. The kernel value represents the inner product between two quantum states $\ket{\phi(\mathbf{x}, \alpha)}$ and $\ket{\phi(\mathbf{x}', \alpha)}$, effectively capturing the state overlap.

Fig.~\ref{fig:state-overlap} illustrates the variation in fidelity overlap for different positions within the input space. This analysis was conducted by selecting a reference point $\mathbf{x}_0 = (0, ..., 0)^\intercal \in \mathbb{R}^w$, where each dimension
is set to zero, aligning with the mean of the standardized time series $\mathbf{s}$. The fidelity overlaps are calculated using $\kappa_\alpha(\mathbf{x}_0, \mathbf{x}) = \left| \braket{\phi(\mathbf{x}_0, \alpha), \phi(\mathbf{x}, \alpha)} \right|^2$, for various points $\mathbf{x}$ from the set $\mathcal{X}_{\text{F}}$. This set comprises points $\mathbf{x}$ with the last $w - 2$ dimensions fixed at zero, and only the first two dimensions varied within the range of the standardized time series' minimum (about -2.7) and maximum values (approximately 2.4).

The figure reveals that points $\mathbf{x}$ closer to the reference point $\mathbf{x}_0$ exhibit higher fidelity values, indicating strong similarity. Conversely, as points deviate further from $\mathbf{x}_0$, their similarity diminishes. This pattern confirms that our quantum kernel effectively functions as expected of a robust kernel function, demonstrating its ability to discern varying degrees of similarity based on proximity in the quantum state space.

\subsubsection{Varying the number of qubits}
\begin{figure}
    \centering
    \subfloat[Number of qubits vs LL.\label{fig:qubits-vs-mae}]{
        \includegraphics[width=0.21\textwidth]{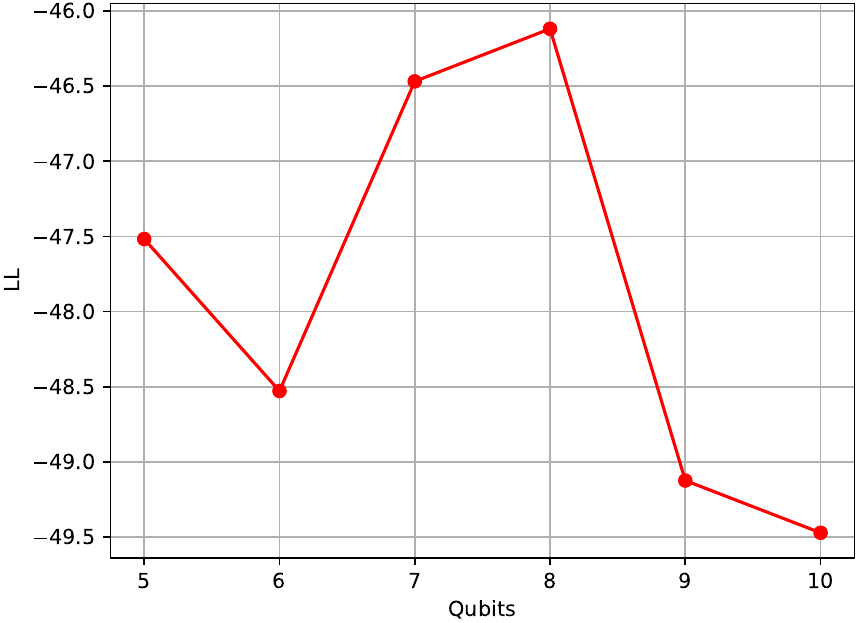}
    }
    \subfloat[Number of qubits vs MAE.\label{fig:qubits-vs-LL}]{
        \includegraphics[width=0.21\textwidth]{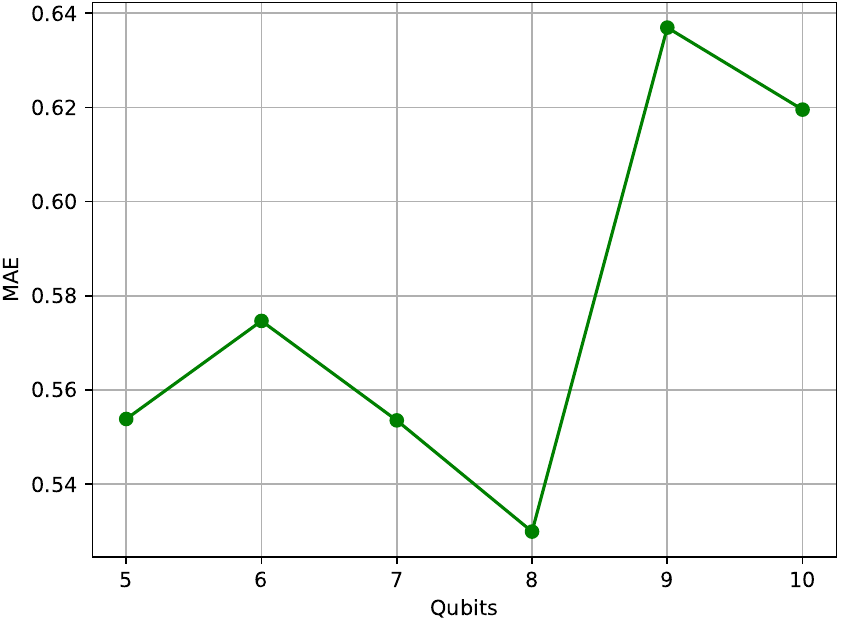}
    }
    
    \caption{The graphs illustrate the impact of the number of qubits on the LL and MAE test metrics, showing optimal performance with 8 qubits and indicating that a higher qubit count can improve results with adequate data.}
    \label{fig:qubits-vs-metrics}
\end{figure}

To assess how the number of qubits, equating to window length, affects our model's performance, we conducted an ablation study focusing on the relationship between the quantity of qubits and various performance metrics. Fig.~\ref{fig:qubits-vs-metrics} illustrates the impact of changing the number of qubits on the LL and MAE during the testing phase. We explored a range of qubit quantities, specifically $\{5, 6, 7, 8, 9, 10\}$, optimizing the model afresh for each configuration before proceeding to measure the respective metrics. Additionally, two modifications were made from the previously mentioned experiment in subsection \ref{subsec:results-qk}: firstly, the overlap between consecutive windows was increased to 4 time steps; secondly, the total count of synthetically generated time steps was set to 480. These adjustments were made to accommodate a larger dataset required by the increased number of qubits.

Our analysis demonstrated that selecting 8 qubits delivered the most favorable outcomes, marking it as the optimum number of qubits for our model configuration. This finding indicates that an increased number of qubits significantly improved the testing metrics compared to the outcomes observed with 5 qubits. However, enhancing the qubit count beyond 8 led to diminishing returns in performance metrics. Such a decline is largely due to the necessity for a more extensive dataset to support precise GPR model inferences with a higher qubit count. Expanding the dataset substantially, though, introduces computational constraints, given that the quantum circuit runs exhibit a quadratic increase in time complexity relative to the size of the dataset. Therefore, the selection of an optimal qubit number should be carefully balanced against the constraints of dataset size and computational capacity.

\section{Conclusion} \label{sec:Conclusion}
In this work, we introduced a hybrid quantum-classical kernelized approach for probabilistic time series forecasting that goes beyond mere value predictions to effectively quantify uncertainties. Leveraging the sophisticated kernel perspective of quantum machine learning models and the robustness of Gaussian process regression, our approach establishes a nexus between quantum models and kernelized probabilistic forecasting. This methodology, incorporating a quantum feature map inspired by Ising interactions, adeptly captures various complex temporal dependencies essential for accurate forecasting. Additionally, we optimize the model's hyperparameters using gradient-free Bayesian optimization, thus sidestepping the computational intensity of gradient descent. Comparative benchmarks against classical kernel models have not only validated our innovative approach but also confirmed its competitive performance within the field.

Looking ahead, this advancement in quantum-enhanced probabilistic forecasting sets the stage for significant future developments. It is crucial to focus on increasing the efficiency of the model and extending its application across diverse datasets to fully exploit its potential. Furthermore, the exploration of quantum feature maps specifically designed to capture complex time series patterns is of great interest for better understanding of the possible enhancements. The focus should be put towards maps that offer a potential quantum advantage that could significantly propel the capabilities of quantum-enhanced time series forecasting. 

\appendices

\section{Bayesian Optimization} \label{app:BO}

In the realm of complex machine learning challenges, parameter tuning emerges as a critical task often reliant on time-intensive evaluations. These evaluations spring from the execution of a black-box objective function $g$, with parameters $\theta$ drawn from a designated parameter space $\Theta$. Due to the substantial resources required for each evaluation, $g$ is characterized as an expensive-to-evaluate function. The objective, therefore, narrows down to identifying an optimal or near-optimal parameter configuration $\theta^* \in \Theta$ that optimizes $g$ through a minimal number of evaluations:
\begin{equation}
    \theta^* = \underset{\theta \in \Theta}{\arg \max} ; g(\theta).
\end{equation}
In such contexts, resorting to gradient-based optimization proves impractical as it typically necessitates a prohibitive number of evaluations to approximate the optimum of $g$ effectively.

Bayesian optimization (BO) presents a strategic framework aimed at minimizing the evaluation count of $g$, all while incorporating historical evaluation data \cite{frazier2018tutorial, garnett2023bayesian}. This method elegantly balances exploration of highly uncertain regions against exploitation in areas showing promising performance. Central to BO are two components: a surrogate model, which provides a probabilistic representation of the objective function, and an acquisition function that strategizes the subsequent sampling point. Initially, BO embarks on a space exploration using a predefined strategy, typically involving $n_0$ points, either through random selection or employing space-filling designs like Sobol sequences \cite{sobol1967distribution}. This phase yields an initial dataset $\mathcal{D}_I = \{(\theta_i, z_i)\}_{1\leq i \leq n_0}$, capturing the outcomes $z_i = g(\theta_i)$ at each $\theta_i$.

\begin{figure}
    \centering
    \subfloat[GP posterior over the objective function.\label{fig:gpei_bo_GP_objective}]{
        \includegraphics[width=0.47\textwidth]{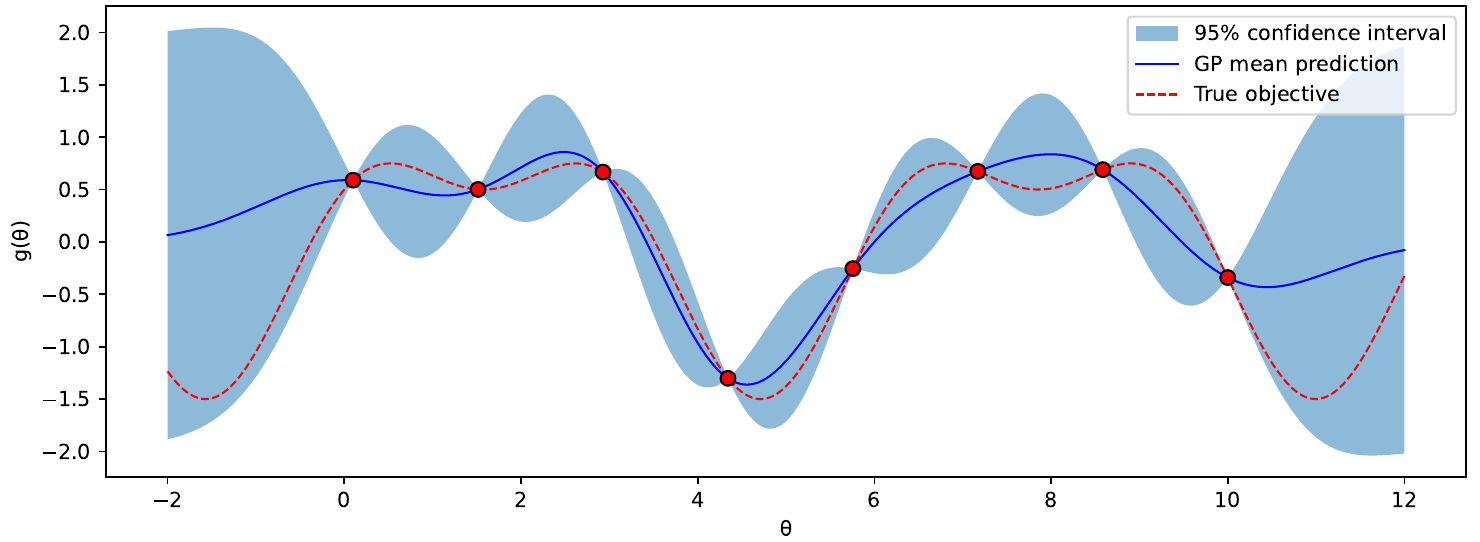}
    }
    \hfill
    \subfloat[Expected improvement acquisition function.\label{fig:gpei_bo_EI}]{
        \includegraphics[width=0.47\textwidth]{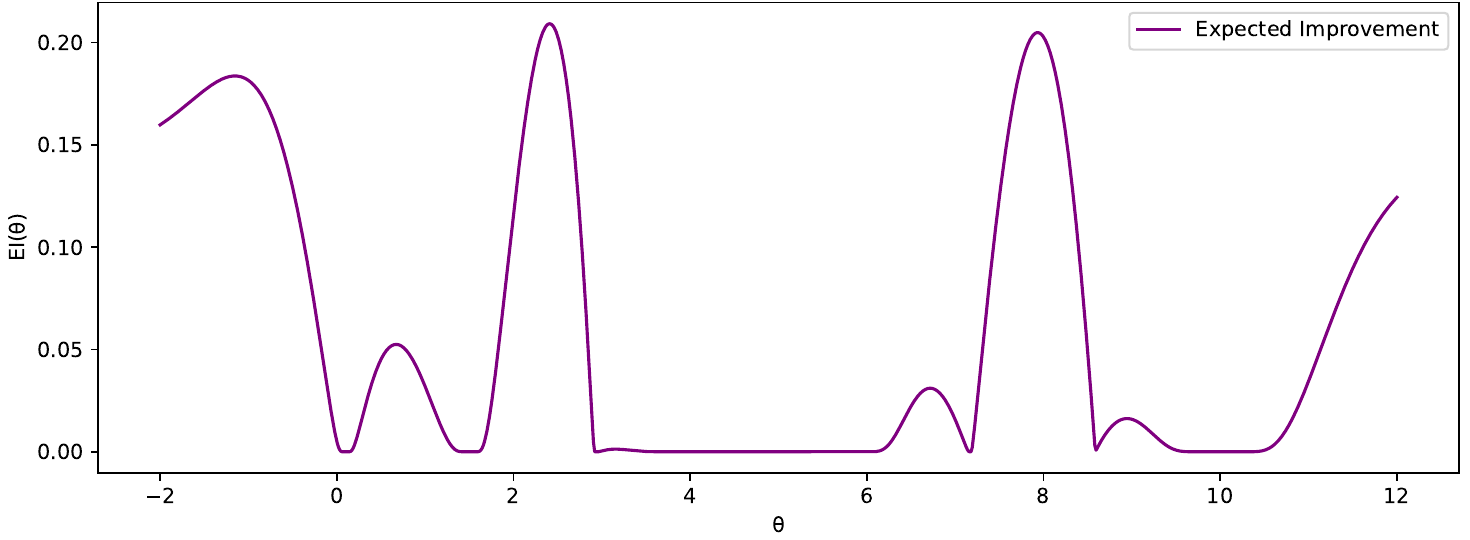}
    }
    \caption{An iteration of the BO process. The top panel shows observations of the objective function with red circles at eight points. The bottom panel show the expected improvement acquisition function that corresponds to this posterior.}
    \label{fig:bo_process}
\end{figure}

In the second step, 
we iteratively refine our understanding of $g$, starting with the initial dataset $\mathcal{D}_I$ to inform a prior over $g$. This process iteratively selects $N$ query points to evaluate, each chosen to maximize the acquisition function, thereby leveraging the posterior distribution of $g$. A Gaussian process is frequently the model of choice for the surrogate, updating the posterior distribution with each new observation and thus enhancing our predictive capability over $g$’s outcomes at untested points. The acquisition function, often chosen to be the expected improvement \cite{frazier2018tutorial}, is defined as:
\begin{equation} \label{eq:EI}
    \text{EI}_{z^*}(\theta) = \mathbb{E}_{g(\theta)}\left[ \max\big(0, g(\theta) - z^*\big)\right],
\end{equation}
where $z^* = \max_i z_i$ denotes the best observed outcome thus far, or the \textit{incumbent}. This function quantifies the utility of sampling $g$ at new candidate points. An illustration of the BO cycle in Fig.~\ref{fig:bo_process} 
showcases how the expected improvement achieves a compromise between areas of high uncertainty and those with favorable observed results.

\section{Further experimental details}
\subsection{Performance metrics}\label{app:performance-metrics}

To perform the quantum-classical comparison, we employed a comprehensive suite of 8 metrics calculated on the test set $\mathcal{D}'$. For each input window in the test set $\mathcal{X}'$, denoted as $\mathbf{x}'_i = (x'_{i,1}, ... x'_{i, w})^\intercal$, we generate a posterior probability distribution $\mathcal{N}_{f'_i} = \mathcal{N}(\bar{f}'_i, \sigma_{f'_i}^2)$, following the approach described in the predictive distribution section \ref{subsec:predictiveDistribution}. These performance metrics are derived from the posterior distributions and the target values $\mathcal{Y}'$, categorized into metrics based on mean predictions and those considering the entire predictive distribution.

The vector of mean predictions for the test set, $\mathbf{f}' = (\bar{f}'_1, ..., \bar{f}'_{c'})^\intercal \in \mathbb{R}^{c'}$, with $c'$ representing the size of $\mathcal{X}'$, is compared against the vector of targets $\mathbf{y}' = (y'_1, ..., y'_{c'})^\intercal$. The first category encompasses traditional error metrics used in the machine learning literature \cite{hodson2022root}, assessing the deviation of mean predictions $\bar{f}'_i$ from the actual targets $y'_i$, including the Mean Squared Error (MSE), the Root Mean Squared Error (RMSE), and the Mean Absolute Error (MAE), defined respectively as
\begin{equation}
    \begin{aligned}
        &\text{MSE}(\mathbf{f}', \mathbf{y}') = \frac{1}{c'} \sum_i (f'_i-y'_i)^2,\\
        &\text{RMSE}(\mathbf{f}', \mathbf{y}') = \sqrt{\text{MSE}(\mathbf{f}', \mathbf{y}')},\\
        &\text{MAE}(\mathbf{f}', \mathbf{y}') = \frac{1}{c'} \sum_i |f'_i-y'_i|.
    \end{aligned}
\end{equation}
Additionally, we calculate another three losses very known in the TSF literature \cite{hyndman2008forecasting}: the Mean Absolute Percentage Error (MAPE), the symmetric MAPE (sMAPE), and the Weighted Average Percentage Error (WAPE). Their computation is achieved as follows:
\begin{equation}
    \begin{aligned}
        &\text{MAPE}(\mathbf{f}', \mathbf{y}') = \frac{1}{c'}\sum_i \left| \frac{y'_i - f'_i}{y'_i} \right|,\\
        &\text{sMAPE}(\mathbf{f}', \mathbf{y}') = \frac{1}{c'}\sum_i \frac{|y'_i - f'_i|}{(y'_i + f'_i)/2},\\
        &\text{WAPE}(\mathbf{f}', \mathbf{y}') = \frac{\sum_i |y'_i - f'_i|}{\sum_i |y'_i|}.
    \end{aligned}
\end{equation}

The second category of metrics we utilized includes probabilistic measures, such as the mean Continuous Ranked Probability Score (mCRPS) and the log likelihood (LL), which evaluate the entire predictive distribution. The CRPS is a measure comparing a single observed value $y$ against a Cumulative Distribution Function (CDF):
\begin{equation}
    \text{CRPS}(F, y) = \int_{-\infty}^{\infty} \left(F(z) - 1_{\{z \geq y\}} \right)^2 dz,
\end{equation}
where the indicator function $1_{{z \geq y}}$ equals 1 when $z \geq y$ and 0 otherwise. For predictions modeled as normal distributions $\mathcal{N}(\bar{f}, \sigma_f^2)$, the CRPS can be determined analytically \cite{taillardat2016calibrated}, facilitating its calculation as:
\begin{equation}
    \begin{aligned}
        \text{CRPS}\left( \mathcal{N}(\bar{f}, \sigma_f^2), y \right) =& \; \sigma_f^2 \bigg( \omega \big(2 . \Phi(\omega) - 1\big)\\
        &+ 2 . \phi(\omega) - \frac{1}{\sqrt{\pi}} \bigg), \\
    \end{aligned}
\end{equation}
where $\Phi(.)$ denotes the cumulative normal distribution function, $\phi(.)$ represents the normal distribution's probability density function, and $\omega = (y - \bar{f}) / \sigma_f$ is the standardized target value. The mCRPS is then the average of the CRPS values calculated for each observation in the test set. 

The log likelihood (LL) metric, on the other hand, quantifies the probability of observing the test set targets $\mathbf{y}'$ given the mean predictions $\mathbf{f}'$, expressed as $\text{LL}(\mathbf{y}', \mathbf{f}') = \log p(\mathbf{y}' | \mathbf{f}')$. This metric, in our context, is given by the following analytical formula:
\begin{equation} \label{eq:log-likelihood}
    \begin{aligned}
        \text{LL}(\mathbf{y}', \mathbf{f}') = \frac{1}{c'}\bigg(&- \frac{1}{2} \sum_{i=1}^{c'} \log(2\pi \sigma_{f'_i}^2) \\
        & - \frac{1}{2 \sigma_{f'_i}^2} \sum_{i=1}^{c'} (\bar{f}'_i - y'_i)^2  \bigg).
    \end{aligned}
\end{equation}
These probabilistic metrics provide a comprehensive evaluation of the model's predictive distribution, offering insights beyond mere point estimates to include uncertainty quantification and the likelihood of observed outcomes.

\subsection{Coding the experiments} \label{app:experiments-implemetation}
In our implementation, we constructed the quantum GPR model leveraging the synergistic capabilities of several libraries, each chosen for their specific strengths in the domain of quantum computing and machine learning optimization. The foundational quantum circuit elements were developed utilizing PennyLane \cite{bergholm2018pennylane}, a library renowned for its seamless integration of quantum computing with machine learning frameworks. This choice allowed us to embed quantum computations directly within our GPR model, paving the way for exploring quantum-enhanced predictive analytics.

For the GP framework, we adopted GPyTorch \cite{gardner2018gpytorch} due to its robust performance in handling large-scale GP models and its efficient computational processes, crucial for the intricate computations our model required. Hyperparameter optimization was conducted using BoTorch \cite{balandat2020botorch}, which offers sophisticated programming tools to efficiently conduct the Bayesian optimization process. The entire optimization workflow was orchestrated through Ax \cite{ax_library}, which provides a structured environment for managing experiments and streamlined the integration with BoTorch. Our source code is publicly accessible
at \href{https://github.com/abdo-aar/quack-tsf}{https://github.com/abdo-aar/quack-tsf}.  

\section*{Acknowledgments}
This work was supported by the Ministère de l’Économie et de l’Innovation du Québec through its contribution to the Quantum AlgoLab of Institut quantique at Université de Sherbrooke. The authors would also like to extend their gratitude to the Natural Sciences and Engineering Research Council of Canada, Prompt Innov, Thales Digital Solutions, and Zetane Systems for their financial support of this research.
\newpage

\bibliographystyle{IEEEtran}
\bibliography{main}

\begin{thebibliography}{10}
\providecommand{\url}[1]{#1}
\csname url@samestyle\endcsname
\providecommand{\newblock}{\relax}
\providecommand{\bibinfo}[2]{#2}
\providecommand{\BIBentrySTDinterwordspacing}{\spaceskip=0pt\relax}
\providecommand{\BIBentryALTinterwordstretchfactor}{4}
\providecommand{\BIBentryALTinterwordspacing}{\spaceskip=\fontdimen2\font plus
\BIBentryALTinterwordstretchfactor\fontdimen3\font minus \fontdimen4\font\relax}
\providecommand{\BIBforeignlanguage}[2]{{%
\expandafter\ifx\csname l@#1\endcsname\relax
\typeout{** WARNING: IEEEtran.bst: No hyphenation pattern has been}%
\typeout{** loaded for the language `#1'. Using the pattern for}%
\typeout{** the default language instead.}%
\else
\language=\csname l@#1\endcsname
\fi
#2}}
\providecommand{\BIBdecl}{\relax}
\BIBdecl

\bibitem{arute2019quantum}
F.~Arute, K.~Arya, R.~Babbush, D.~Bacon, J.~C. Bardin, R.~Barends, R.~Biswas, S.~Boixo, F.~G. Brandao, D.~A. Buell \emph{et~al.}, ``Quantum supremacy using a programmable superconducting processor,'' \emph{Nature}, vol. 574, no. 7779, pp. 505--510, 2019.

\bibitem{bravyi2022future}
S.~Bravyi, O.~Dial, J.~M. Gambetta, D.~Gil, and Z.~Nazario, ``The future of quantum computing with superconducting qubits,'' \emph{Journal of Applied Physics}, vol. 132, no.~16, 2022.

\bibitem{bayerstadler2021industry}
A.~Bayerstadler, G.~Becquin, J.~Binder, T.~Botter, H.~Ehm, T.~Ehmer, M.~Erdmann, N.~Gaus, P.~Harbach, M.~Hess \emph{et~al.}, ``Industry quantum computing applications,'' \emph{EPJ Quantum Technology}, vol.~8, no.~1, p.~25, 2021.

\bibitem{bova2021commercial}
F.~Bova, A.~Goldfarb, and R.~G. Melko, ``Commercial applications of quantum computing,'' \emph{EPJ quantum technology}, vol.~8, no.~1, p.~2, 2021.

\bibitem{bornens2023variational}
A.-S. Bornens and M.~Nowak, ``Variational quantum algorithms on cat qubits,'' \emph{arXiv preprint arXiv:2305.14143}, 2023.

\bibitem{madsen2022quantum}
L.~S. Madsen, F.~Laudenbach, M.~F. Askarani, F.~Rortais, T.~Vincent, J.~F. Bulmer, F.~M. Miatto, L.~Neuhaus, L.~G. Helt, M.~J. Collins \emph{et~al.}, ``Quantum computational advantage with a programmable photonic processor,'' \emph{Nature}, vol. 606, no. 7912, pp. 75--81, 2022.

\bibitem{blais2021circuit}
A.~Blais, A.~L. Grimsmo, S.~M. Girvin, and A.~Wallraff, ``Circuit quantum electrodynamics,'' \emph{Reviews of Modern Physics}, vol.~93, no.~2, p. 025005, 2021.

\bibitem{kim2023evidence}
Y.~Kim, A.~Eddins, S.~Anand, K.~X. Wei, E.~Van Den~Berg, S.~Rosenblatt, H.~Nayfeh, Y.~Wu, M.~Zaletel, K.~Temme \emph{et~al.}, ``Evidence for the utility of quantum computing before fault tolerance,'' \emph{Nature}, vol. 618, no. 7965, pp. 500--505, 2023.

\bibitem{bravyi2020quantum}
S.~Bravyi, D.~Gosset, R.~K{\"o}nig, and M.~Tomamichel, ``Quantum advantage with noisy shallow circuits,'' \emph{Nature Physics}, vol.~16, no.~10, pp. 1040--1045, 2020.

\bibitem{rainjonneau2023quantum}
S.~Rainjonneau, I.~Tokarev, S.~Iudin, S.~Rayaprolu, K.~Pinto, D.~Lemtiuzhnikova, M.~Koblan, E.~Barashov, M.~Kordzanganeh, M.~Pflitsch \emph{et~al.}, ``Quantum algorithms applied to satellite mission planning for earth observation,'' \emph{IEEE Journal of Selected Topics in Applied Earth Observations and Remote Sensing}, 2023.

\bibitem{piatkowski2022towards}
N.~Piatkowski, T.~Gerlach, R.~Hugues, R.~Sifa, C.~Bauckhage, and F.~Barbaresco, ``Towards bundle adjustment for satellite imaging via quantum machine learning,'' in \emph{2022 25th International Conference on Information Fusion (FUSION)}.\hskip 1em plus 0.5em minus 0.4em\relax IEEE, 2022, pp. 1--8.

\bibitem{liu2021rigorous}
Y.~Liu, S.~Arunachalam, and K.~Temme, ``A rigorous and robust quantum speed-up in supervised machine learning,'' \emph{Nature Physics}, vol.~17, no.~9, pp. 1013--1017, 2021.

\bibitem{preskill2018quantum}
J.~Preskill, ``Quantum computing in the nisq era and beyond,'' \emph{Quantum}, vol.~2, p.~79, 2018.

\bibitem{havlivcek2019supervised}
V.~Havl{\'\i}{\v{c}}ek, A.~D. C{\'o}rcoles, K.~Temme, A.~W. Harrow, A.~Kandala, J.~M. Chow, and J.~M. Gambetta, ``Supervised learning with quantum-enhanced feature spaces,'' \emph{Nature}, vol. 567, no. 7747, pp. 209--212, 2019.

\bibitem{schuld2021supervised}
M.~Schuld, ``Supervised quantum machine learning models are kernel methods,'' \emph{arXiv preprint arXiv:2101.11020}, 2021.

\bibitem{scholkopf2001generalized}
B.~Sch{\"o}lkopf, R.~Herbrich, and A.~J. Smola, ``A generalized representer theorem,'' in \emph{International conference on computational learning theory}.\hskip 1em plus 0.5em minus 0.4em\relax Springer, 2001, pp. 416--426.

\bibitem{burges1998tutorial}
C.~J. Burges, ``A tutorial on support vector machines for pattern recognition,'' \emph{Data mining and knowledge discovery}, vol.~2, no.~2, pp. 121--167, 1998.

\bibitem{vovk2013kernel}
V.~Vovk, ``Kernel ridge regression,'' in \emph{Empirical Inference: Festschrift in Honor of Vladimir N. Vapnik}.\hskip 1em plus 0.5em minus 0.4em\relax Springer, 2013, pp. 105--116.

\bibitem{huang2022quantum}
H.-Y. Huang, M.~Broughton, J.~Cotler, S.~Chen, J.~Li, M.~Mohseni, H.~Neven, R.~Babbush, R.~Kueng, J.~Preskill \emph{et~al.}, ``Quantum advantage in learning from experiments,'' \emph{Science}, vol. 376, no. 6598, pp. 1182--1186, 2022.

\bibitem{gui2014financial}
B.~Gui, X.~Wei, Q.~Shen, J.~Qi, and L.~Guo, ``Financial time series forecasting using support vector machine,'' in \emph{2014 Tenth International Conference on Computational Intelligence and Security}.\hskip 1em plus 0.5em minus 0.4em\relax IEEE, 2014, pp. 39--43.

\bibitem{thakkar2024improved}
S.~Thakkar, S.~Kazdaghli, N.~Mathur, I.~Kerenidis, A.~J. Ferreira-Martins, and S.~Brito, ``Improved financial forecasting via quantum machine learning,'' \emph{Quantum Machine Intelligence}, vol.~6, no.~1, p.~27, 2024.

\bibitem{kalfon2023successive}
B.~Kalfon, S.~Cherkaoui, J.-F. Laprade, O.~Ahmad, and S.~Wang, ``Successive data injection in conditional quantum gan applied to time series anomaly detection,'' \emph{IET Quantum Communication}, 2023.

\bibitem{liu2018quantum}
N.~Liu and P.~Rebentrost, ``Quantum machine learning for quantum anomaly detection,'' \emph{Physical Review A}, vol.~97, no.~4, p. 042315, 2018.

\bibitem{aaraba2024fr}
A.~Aaraba, S.~Wang, and J.-M. Patenaude, ``\uppercase{FR\textsuperscript{3}LS}: A forecasting model with robust and reduced redundancy latent series,'' in \emph{Pacific-Asia Conference on Knowledge Discovery and Data Mining}.\hskip 1em plus 0.5em minus 0.4em\relax Springer, 2024, pp. 3--15.

\bibitem{rasmussen2006gaussian}
C.~E. Rasmussen, C.~K. Williams \emph{et~al.}, \emph{Gaussian processes for machine learning}.\hskip 1em plus 0.5em minus 0.4em\relax Springer, 2006, vol.~1.

\bibitem{rebentrost2014quantum}
P.~Rebentrost, M.~Mohseni, and S.~Lloyd, ``Quantum support vector machine for big data classification,'' \emph{Physical review letters}, vol. 113, no.~13, p. 130503, 2014.

\bibitem{rapp2024quantum}
F.~Rapp and M.~Roth, ``Quantum gaussian process regression for bayesian optimization,'' \emph{Quantum Machine Intelligence}, vol.~6, no.~1, p.~5, 2024.

\bibitem{dai2022quantum}
J.~Dai and R.~V. Krems, ``Quantum gaussian process model of potential energy surface for a polyatomic molecule,'' \emph{The Journal of Chemical Physics}, vol. 156, no.~18, 2022.

\bibitem{cerezo2021variational}
M.~Cerezo, A.~Arrasmith, R.~Babbush, S.~C. Benjamin, S.~Endo, K.~Fujii, J.~R. McClean, K.~Mitarai, X.~Yuan, L.~Cincio \emph{et~al.}, ``Variational quantum algorithms,'' \emph{Nature Reviews Physics}, vol.~3, no.~9, pp. 625--644, 2021.

\bibitem{schuld2021machine}
M.~Schuld and F.~Petruccione, \emph{Machine learning with quantum computers}.\hskip 1em plus 0.5em minus 0.4em\relax Springer, 2021.

\bibitem{zeguendry2023quantum}
A.~Zeguendry, Z.~Jarir, and M.~Quafafou, ``Quantum machine learning: A review and case studies,'' \emph{Entropy}, vol.~25, no.~2, p. 287, 2023.

\bibitem{baker2023massively}
J.~S. Baker, G.~Park, K.~Yu, A.~Ghukasyan, O.~Goktas, and S.~K. Radha, ``Massively parallel hybrid quantum-classical machine learning for kernelized time-series classification,'' \emph{arXiv preprint arXiv:2305.05881}, 2023.

\bibitem{canatar2022bandwidth}
A.~Canatar, E.~Peters, C.~Pehlevan, S.~M. Wild, and R.~Shaydulin, ``Bandwidth enables generalization in quantum kernel models,'' \emph{arXiv preprint arXiv:2206.06686}, 2022.

\bibitem{shaydulin2022importance}
R.~Shaydulin and S.~M. Wild, ``Importance of kernel bandwidth in quantum machine learning,'' \emph{Physical Review A}, vol. 106, no.~4, p. 042407, 2022.

\bibitem{mitarai2018quantum}
K.~Mitarai, M.~Negoro, M.~Kitagawa, and K.~Fujii, ``Quantum circuit learning,'' \emph{Physical Review A}, vol.~98, no.~3, p. 032309, 2018.

\bibitem{ament2024unexpected}
S.~Ament, S.~Daulton, D.~Eriksson, M.~Balandat, and E.~Bakshy, ``Unexpected improvements to expected improvement for bayesian optimization,'' \emph{Advances in Neural Information Processing Systems}, vol.~36, 2024.

\bibitem{byrd1995limited}
R.~H. Byrd, P.~Lu, J.~Nocedal, and C.~Zhu, ``A limited memory algorithm for bound constrained optimization,'' \emph{SIAM Journal on scientific computing}, vol.~16, no.~5, pp. 1190--1208, 1995.

\bibitem{zhu1997algorithm}
C.~Zhu, R.~H. Byrd, P.~Lu, and J.~Nocedal, ``Algorithm 778: L-bfgs-b: Fortran subroutines for large-scale bound-constrained optimization,'' \emph{ACM Transactions on mathematical software (TOMS)}, vol.~23, no.~4, pp. 550--560, 1997.

\bibitem{huang2021mcclean}
B.~Huang, B.~Mohseni, and N.~Boixo, ``Mcclean: Power of data in quantum machine learning,'' \emph{Nature Communications}, vol.~12, no. 2631, pp. 2041--1723, 2021.

\bibitem{vert2004primer}
J.-P. Vert, K.~Tsuda, and B.~Sch{\"o}lkopf, ``A primer on kernel methods,'' 2004.

\bibitem{genton2001classes}
M.~G. Genton, ``Classes of kernels for machine learning: a statistics perspective,'' \emph{Journal of machine learning research}, vol.~2, no. Dec, pp. 299--312, 2001.

\bibitem{duvenaud2014kernel}
D.~Duvenaud, ``The kernel cookbook: Advice on covariance functions,'' \emph{URL https://www. cs. toronto. edu/duvenaud/cookbook}, 2014.

\bibitem{mackay1998introduction}
D.~J. MacKay \emph{et~al.}, ``Introduction to gaussian processes,'' \emph{NATO ASI series F computer and systems sciences}, vol. 168, pp. 133--166, 1998.

\bibitem{frazier2018tutorial}
P.~I. Frazier, ``A tutorial on bayesian optimization,'' \emph{arXiv preprint arXiv:1807.02811}, 2018.

\bibitem{garnett2023bayesian}
R.~Garnett, \emph{Bayesian optimization}.\hskip 1em plus 0.5em minus 0.4em\relax Cambridge University Press, 2023.

\bibitem{sobol1967distribution}
I.~M. Sobol', ``On the distribution of points in a cube and the approximate evaluation of integrals,'' \emph{Zhurnal Vychislitel'noi Matematiki i Matematicheskoi Fiziki}, vol.~7, no.~4, pp. 784--802, 1967.

\bibitem{hodson2022root}
T.~O. Hodson, ``Root mean square error (rmse) or mean absolute error (mae): When to use them or not,'' \emph{Geoscientific Model Development Discussions}, vol. 2022, pp. 1--10, 2022.

\bibitem{hyndman2008forecasting}
R.~Hyndman, A.~B. Koehler, J.~K. Ord, and R.~D. Snyder, \emph{Forecasting with exponential smoothing: the state space approach}.\hskip 1em plus 0.5em minus 0.4em\relax Springer Science \& Business Media, 2008.

\bibitem{taillardat2016calibrated}
M.~Taillardat, O.~Mestre, M.~Zamo, and P.~Naveau, ``Calibrated ensemble forecasts using quantile regression forests and ensemble model output statistics,'' \emph{Monthly Weather Review}, vol. 144, no.~6, pp. 2375--2393, 2016.

\bibitem{bergholm2018pennylane}
V.~Bergholm, J.~Izaac, M.~Schuld, C.~Gogolin, S.~Ahmed, V.~Ajith, M.~S. Alam, G.~Alonso-Linaje, B.~AkashNarayanan, A.~Asadi \emph{et~al.}, ``Pennylane: Automatic differentiation of hybrid quantum-classical computations,'' \emph{arXiv preprint arXiv:1811.04968}, 2018.

\bibitem{gardner2018gpytorch}
J.~Gardner, G.~Pleiss, K.~Q. Weinberger, D.~Bindel, and A.~G. Wilson, ``Gpytorch: Blackbox matrix-matrix gaussian process inference with gpu acceleration,'' \emph{Advances in neural information processing systems}, vol.~31, 2018.

\bibitem{balandat2020botorch}
M.~Balandat, B.~Karrer, D.~Jiang, S.~Daulton, B.~Letham, A.~G. Wilson, and E.~Bakshy, ``Botorch: A framework for efficient monte-carlo bayesian optimization,'' \emph{Advances in neural information processing systems}, vol.~33, pp. 21\,524--21\,538, 2020.

\bibitem{ax_library}
\BIBentryALTinterwordspacing
``Ax: Adaptive experimentaion platform,'' 2022. [Online]. Available: \url{https://ax.dev/}
\BIBentrySTDinterwordspacing

\end{thebibliography}

\end{document}